\documentclass[10pt,journal,compsoc]{IEEEtran}

\ifCLASSOPTIONcompsoc
  \usepackage[nocompress]{cite}
  \usepackage[caption=false,font=footnotesize,labelfont=sf,textfont=sf]{subfig}
\else
  \usepackage{cite}
  \usepackage[caption=false,font=footnotesize]{subfig}
\fi
\ifCLASSINFOpdf
  \usepackage[pdftex]{graphicx}
  \DeclareGraphicsExtensions{.pdf,.jpeg,.png}
\else
  \usepackage[dvips]{graphicx}
  \DeclareGraphicsExtensions{.eps}
\fi
\usepackage{amsmath,amssymb,amsfonts,dsfont,bm,mathrsfs}
\usepackage{algorithmic}
\usepackage{booktabs,multirow,adjustbox}
\usepackage{float}
\usepackage[table]{xcolor}
\usepackage{collcell}
\usepackage{xifthen}
\usepackage{pgf}
\usepackage[pagebackref=false,breaklinks=true,colorlinks=true,citecolor=blue,bookmarks=false]{hyperref}
\usepackage{xcolor}
\usepackage{soul}
\usepackage[capitalize]{cleveref} 
\usepackage{graphicx}

\def\ColorMode{hsb}
\newcommand{\ColCell}[1]{
  \ifthenelse{\isempty{#1}}{}{
    \pgfmathparse{#1<2?1:0}                          
      \ifnum\pgfmathresult=0\relax\color{white}\fi
    \pgfmathsetmacro\compA{0}                        
    \pgfmathsetmacro\compB{(abs(#1)==1?0:abs(#1))/1} 
    \pgfmathsetmacro\compC{1}                        
    \edef\x{\noexpand\centering\noexpand\cellcolor[\ColorMode]{\compA,\compB,\compC}}\x #1
  }
}
\newcolumntype{C}[1]{>{\collectcell\ColCell}m{#1}<{\endcollectcell}}  

\newcommand{\x}{{\rm\bf x}}      

\crefname{section}{Sec.}{Secs.}
\Crefname{section}{Section}{Sections}
\crefname{table}{Tab.}{Tabs.}
\Crefname{table}{Table}{Tables}
\crefname{figure}{Fig.}{Figs.}
\Crefname{figure}{Figure}{Figures}
\crefname{equation}{Eq.}{Eqs.}
\Crefname{equation}{Equation}{Equations}

\newcommand{\tofill}[1]{\textcolor{red}{[TBD]}}

\begin{document}


\newcommand{\titlename}{Spatial Steerability of GANs via Self-Supervision from Discriminator}
\title{\titlename}



\author{
Jianyuan Wang$^*$,
Lalit Bhagat$^*$,
Ceyuan Yang, 
Yinghao Xu,
Yujun Shen,
Hongdong Li,
and Bolei Zhou\\
   \IEEEcompsocitemizethanks{
   \IEEEcompsocthanksitem J. Wang is with the University of Oxford, Oxford, United Kingdom.\protect
   \IEEEcompsocthanksitem L. Bhagat and B. Zhou are with the Computer Science Department, University of California, Los Angeles, California, United States.\protect
   \IEEEcompsocthanksitem C. Yang and Y. Xu are with the Department of Information Engineering,  Chinese University of Hong Kong, Hong Kong SAR, China.\protect
   \IEEEcompsocthanksitem Y. Shen is with Ant Group, China.\protect
   \IEEEcompsocthanksitem H. Li is with the College of Engineering and Computer Science, the Australian National University, Canberra, Australia.\protect
   \IEEEcompsocthanksitem $^*$ denotes equal contribution.\protect
   
   }%
}


\IEEEtitleabstractindextext{

\begin{abstract}

Generative models make huge progress to the photorealistic image synthesis in recent years. 
%
To enable human to steer the image generation process and customize the output, many works explore the interpretable dimensions of the latent space in GANs.
Existing methods edit the attributes of the output image such as orientation or color scheme by varying the latent code along certain directions.
%
However, these methods usually require additional human annotations for each pretrained model, and they mostly focus on editing global attributes.
%
In this work, we propose a self-supervised approach to improve the spatial steerability of GANs without searching for steerable directions in the latent space or requiring extra annotations.
Specifically, we design randomly sampled Gaussian heatmaps to be encoded into the intermediate layers of generative models as spatial inductive bias. Along with training the GAN model from scratch, these heatmaps are being aligned with the emerging attention of the GAN's discriminator in a self-supervised learning manner. 
%
During inference, users can interact with the spatial heatmaps in an intuitive manner, enabling them to edit the output image by adjusting the scene layout, moving, or removing objects. Moreover, we incorporate DragGAN into our framework, which facilitates fine-grained manipulation within a reasonable time and supports a coarse-to-fine editing process. 
%
Extensive experiments show that the proposed method not only enables spatial editing over human faces, animal faces, outdoor scenes, and complicated multi-object indoor scenes but also brings improvement in synthesis quality.
%
%
Code, models, and demo video are available at \protect\url{https://genforce.github.io/SpatialGAN/}
\end{abstract}


}

\maketitle
\IEEEdisplaynontitleabstractindextext
\IEEEpeerreviewmaketitle

\section{Introduction}\label{sec:intro}
\begin{figure*}[htbp]
    \centering
    \includegraphics[width=\linewidth]{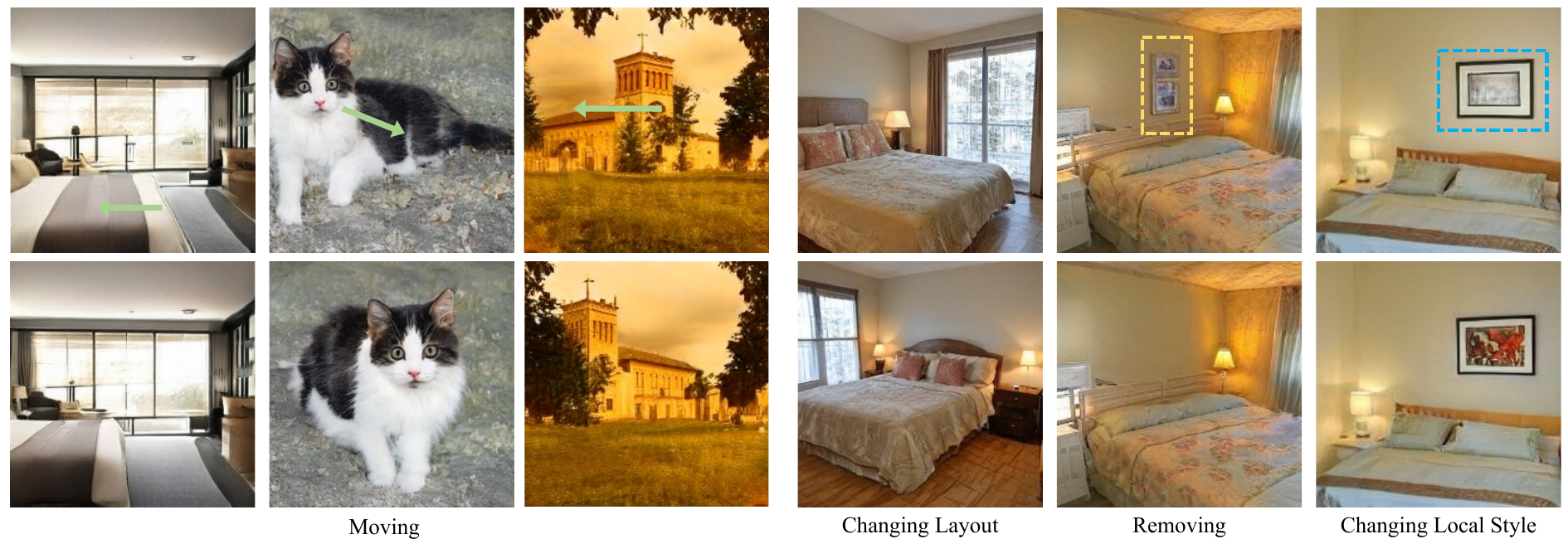}
    \caption{\textbf{Illustration of Spatial Manipulations.} Our method enables various spatial manipulations for image generation, like moving a bed, a cat, or a building (green arrow), controlling the image layout, removing a drawing (yellow box), and changing the local appearance (blue box). 
    }
    \label{Fig:Manipulation}
\end{figure*}

\IEEEPARstart{G}{enerative} Adversarial Network (GAN) has made huge progress to high-quality image synthesis~\cite{goodfellow2014generative, dcgan, bigGAN, stylegan, stylegan2}.
%
%
GAN is formulated as a two-player game between a generator ($G$) and a discriminator ($D$)~\cite{goodfellow2014generative}, where $G$ maps a random distribution to real-world observation, and $D$ competes with $G$ by distinguishing the generated images from the real ones.
It is found that the latent space of $G$ contains disentangled subspaces, which align with various image attributes, \textit{e.g.}, the age of human faces~\cite{shen2020interpreting}, the layout of indoor scenes~\cite{higan}, and the pose of vehicles~\cite{shen2021closed}.
Researchers utilize such properties to study the knowledge learned by GANs and facilitate interactive editing over the output image.

However, the existing methods~\cite{goetschalckx2019ganalyze,shen2020interpreting,voynov2020unsupervised,yang2021semantic,harkonen2020ganspace,shen2021closed} mostly require additional information like extra annotation or human selection. 
For a target attribute and a given generator, they search for an associated direction in the high-dimensional latent space and then change the image attribute via varying the latent code along the found direction.
A typical approach is to first sample numerous images from the latent space, label them regarding the target attribute, and then learn to find the tangent direction, which could be expensive, unstable, and sometimes inapplicable.
%
%
Some recent works~\cite{shen2021closed,harkonen2020ganspace,voynov2020unsupervised} identify the essential directions in the latent space via the techniques like Principal Component Analysis (PCA).
Unfortunately, these methods cannot guarantee which attributes to be found, and human still needs to distinguish which attribute each direction corresponds to and select the meaningful ones. Moreover, the spatial steerability of generative models, such as moving an object or changing the local appearance of an object in the output image, is much less explored. 

In this work, we propose a novel self-supervision approach called \textit{SpatialGAN} to achieve spatial steerability of GANs without searching for steerable directions in the latent space.
%
%
%
It allows human users to perform various spatial manipulations in the image generation, such as moving an object and removing an object in a scene, changing the style of a region, or globally controlling the structure/layout of an image.
Some examples are shown in \cref{Fig:Manipulation}.
Previous work shows that the class specific attention maps emerge in image classification networks~\cite{cam}. We reveal that the discriminator of GAN, as a bi-classifier for adversarial training, also has emerging attention highlighting the informative region of the synthesized image.
%
Therefore, we incorporate a design of spatial heatmaps as inductive bias in the generator, and then learn to align them with the attention maps from the discriminator in a self-supervised learning manner.
Specifically, we randomly sample heatmaps and encode them into the intermediate layers of $G$ to guide its spatial focus. 
%
%
To ensure the encoded heatmaps focusing on the meaningful regions of the synthesized image, we regularize the generator's heatmap to be aligned with the discriminator's attention map on the synthesized image. 
In other words, we utilize the attention map emerging from the discriminator to guide the heatmap in the generator. 
%
The whole process follows a self-supervised learning manner and does not involve extra annotation or statistical information.
It trains the generator to synthesize the image based on the input heatmaps, and improves the spatial steerability of the model, \textit{i.e.}, we could edit the heatmap to spatially control the output synthesis during inference. 
%
The preliminary result of such spatial steerability was shown at our conference version~\cite{Wang_2022_CVPR}, where the main focus was to improve synthesis quality by incorporating heatmaps as an inductive bias, and the steerability is a byproduct. 
We initially demonstrated this ability in single-object scenes, which sharply focus on one primary element. 
This could range from artifacts, human faces, animals, to more contextual settings like a lone car on a street or an isolated historical building. Such scenes, due to their focus on a singular primary element, are relatively straightforward to analyze and interpret.
%
%
Here, moving to indoor scenes, we address the complexities of multi-object indoor scenes. 
For instance, a living room scene might include a sofa, a coffee table, and artwork on the walls. Each element contributes to the scene's overall composition, requiring careful placement to ensure a realistic representation. 
These scenes pose a notable challenge due to their multiple points of focus; a single point on the heatmap is insufficient for capturing the scene's full dynamics. To accurately generate such scenes, it is imperative to understand not only each object individually but also how they collectively interact within the space. 

%
%
Compared to the conference version, this journal paper has achieved notable advancements in spatial steerability. 
Specifically, (i) to enable the spatial steerability in complex indoor scenes with multiple objects, we have developed a new heatmap construction strategy, encoding method, and self-supervision training objective; 
(ii) our enhanced method facilitates more sophisticated spatial manipulations, such as removing objects and changing the style of a local region, as illustrated \cref{Fig:Manipulation}; 
(iii) we have also significantly improved the synthesis quality of single-object scenes using a refined heatmap processing strategy; 
(iv) we integrate the recent progress in point-based manipulation (e.g., DragGAN\cite{pan2023drag}) into our method. This integration combines the strengths of both approaches to achieve high-quality, fine-grained manipulation in a reasonable time. Our results not only demonstrate the individual merits of our framework but also its complementary relationship with DragGAN, highlighting the versatility and effectiveness of our approach;  
(v) we develop a new user interface to illustrate our manipulation ability; 
and (vi) we present an expanded set of results and provide a comprehensive analysis that highlights significant advancements in both manipulation capability and synthesis quality.  

\section{Related Work}
\label{sec:related}

\subsection{Generative Adversarial Networks} GANs~\cite{goodfellow2014generative} achieve great success in photorealistic image generation.
It aims to learn the target distribution via a minimax two-player game of generator and discriminator.
The generator usually takes in a random latent code and produces a synthesis image.
Researchers have developed numerous techniques to improve the synthesis quality of GANs, through a Laplacian pyramid framework~\cite{denton2015deep}, an all-convolutional deep neural network~\cite{dcgan}, progressive training~\cite{pggan}, spectral normalization~\cite{miyato2018spectral,zhang2019self}, and large-sacle GAN training~\cite{bigGAN,BigBiGAN}.
Some methods also incorporate additional information into discriminator or generator, such as pixel-wise representation~\cite{unetgan}, 3D pose~\cite{giraffe}, or neighboring instances~\cite{instanceGAN}.
In recent years, the style-based architecture StyleGAN~\cite{stylegan} and StyleGAN2~\cite{stylegan2} have become the state-of-the-art method for image synthesis, by separating high-level attributes. \\
%
%

\subsection{GAN Manipulation}
To understand the generation process of GANs and support human customization of the output image, researchers have been trying to control the output synthesis.
A popular way is to leverage the rich semantic information in the latent space of GAN.
They identify steerable properties as some directions in the latent space and vary the latent code accordingly.
For a certain attribute, they search for a certain direction in the latent space, and then alter the target attribute via moving the latent code z along the searched direction~\cite{goetschalckx2019ganalyze,shen2020interpreting,voynov2020unsupervised,yang2021semantic,harkonen2020ganspace}.
However, for each pre-trained GAN model (\textit{i.e.}, pre-trained latent space), these methods require to annotate a collection of the generated samples to train linear classifiers in the latent space~\cite{goetschalckx2019ganalyze,higan,shen2020interpreting} or utilize sample statistics~\cite{Plumerault2020Controlling, Jahanian2020On}. %
These requirements are expensive and it can only accommodate limited number of attributes.
Some recent works~\cite{shen2021closed,harkonen2020ganspace,voynov2020unsupervised,zhu2022resefa} search for steerable directions using techniques like Principal Component Analysis (PCA) in an unsupervised manner.
Unfortunately, it does not guarantee the attributes of found directions. 
For example, in order to achieve spatial control of the image generation, the user has to manually check the effect of found directions, while the one corresponding to the desired spatial manipulation may not exist.
Instead, our proposed method brings spatial steerability of GANs without searching for steerable directions in the latent space, which avoids the requirements of extra annotation or human selection.
%
A concurrent work \cite{epstein2022blobgan} achieves similar capabilities by learning a category-specific middle-level representation. However, this method necessitates a significantly larger amount of training data and computational resources compared to our approach.
A recent innovation, DragGAN \cite{pan2023drag}, introduces a point-based manipulation technique that allows for fine editing by performing optimization on the latent code during inference. 
While this method does provide granular control, it is constrained by the time-intensive nature of the optimization process, particularly when broader, more coarse adjustments are required. In contrast, our method facilitates the movement of objects seamlessly and intuitively without resorting to any form of optimization, thereby offering a more efficient alternative for real-time spatial editing in generative models

\section{Method}\label{sec:method}

\begin{figure*}[t]
    \centering
    \includegraphics[width=\linewidth]{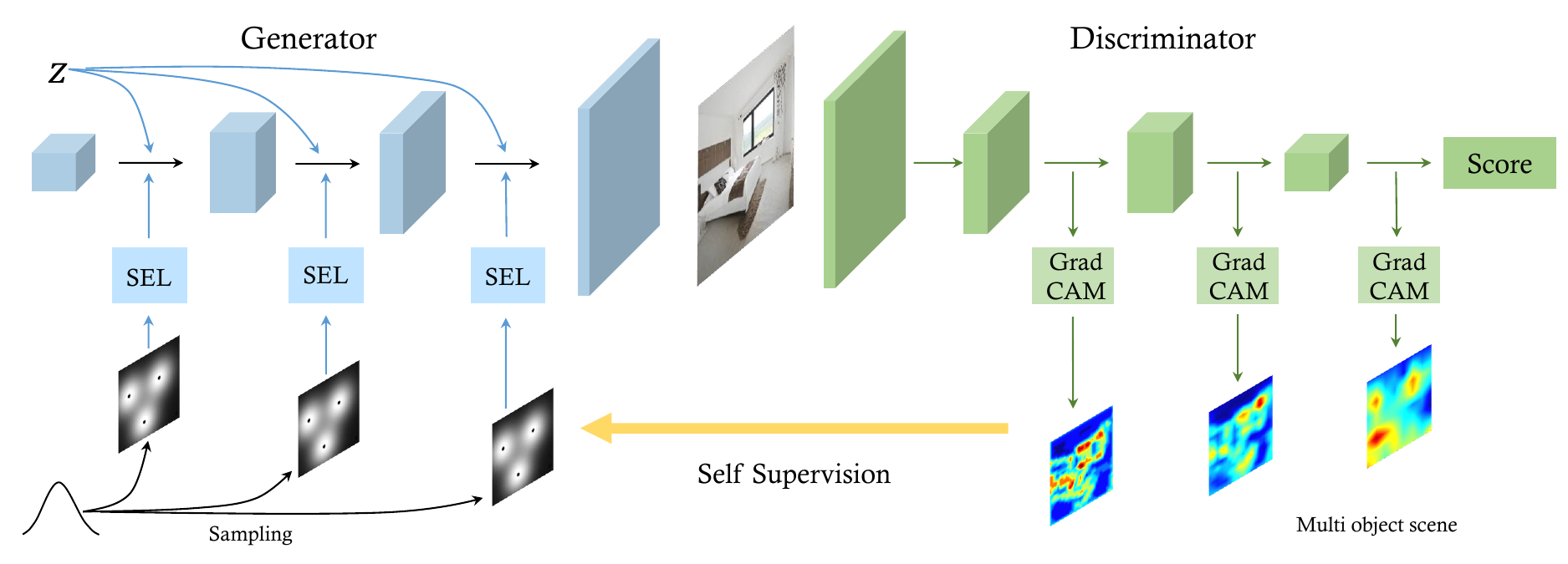}
    \vspace{-5pt}
    \caption{\textbf{Illustration of SpatialGAN.} We conduct spatial encoding in $G$ and align its spatial awareness with $D$ attention maps. Specifically, we randomly sample spatial heatmaps and encode them into $G$ via the spatial encoding layer (SEL). To implement the alignment during training, we calculate $D$ attention maps over the generated samples via GradCAM. 
    }
    \label{Fig:Landscape}
\end{figure*}

In this section, we first analyze the spatial attention of the GAN's discriminator in \cref{subsec:datt}, which serves as the guidance for implementing the pseudo attention mechanism in the generator. 
We then introduce the hierarchical heatmap sampling strategy and the heatmap encoding methods in \cref{subsec:spatial-awareness-to-g}.
%
%
Compared to our conference version, we find that a heatmap with too fine-grained scale may not benefit the image synthesis, and hence proposes a coarse processing of heatmap  to improve the synthesis quality. 
%
%
Furthermore, to address the complexities of indoor scenes with multiple objects, we update the method of heatmap sampling and heatmap encoding.
In \cref{subsec:feedback-mechanism}, we discuss utilizing the emerging attention map from the discriminator as a self-supervision signal for image synthesis which paves the way for spatial editing. 
The overall framework is illustrated in \cref{Fig:Landscape},  primarily involving two steps: the explicit encoding of spatial inductive bias into $G$ and using the emerging attention map from $D$ to supervise $G$. 
Different from the conference version, we also introduce a new self-supervision objective tailored for intricate indoor scenes, enabling advanced spatial manipulations.
Lastly, in \cref{subsec:DragGAN-x-sGAN}, we extend our discussion to the integration of our SpatialGAN  with DragGAN~\cite{pan2023drag}, a point-based image manipulation technique. This synergy leverages the strengths of both approaches, facilitating more efficient and flexible manipulations in generative models.

\subsection{Spatial Attention of Discriminator}\label{subsec:datt}

We first investigate the behavior of $D$ in the spatial domain, because $D$ is designed to differentiate between the real distribution and the distribution generated by $G$, \textit{i.e.}, $D$ is an adversary as well as acting as a teacher for $G$ in this two-player game.

Prior work on network interpretability, like CAM~\cite{cam}, has found that a classification network tends to focus on some discriminative spatial regions to categorize a given image.
However, the discriminator in GANs is trained with the relatively weak supervision, \textit{i.e.}, only having real or fake labels.
Whether it can learn the attentive property from such a bi-classification task remains unknown.
To look under the hood, we apply GradCAM~\cite{gradcam} as an interpretability tool on the well-trained discriminator of StyleGAN2~\cite{stylegan2}. 

Specifically, for a certain layer and a certain class, GradCAM calculates the importance weight of each neuron by average-pooling the gradients back-propagated from the final classification score, over the width and height. 
It then computes the attention map as a weighted combination of the importance weight and the forward activation maps, followed by a ReLU~\cite{glorot2011deep} activation.
The attention map has the same spatial shape as the corresponding feature map.
In this work, we report the GradCAM attention maps all using gradients computed via maximizing the output of $D$.
It reflects the spatial preference of $D$ in making a `real' decision.
In practice we find the attention maps are almost the same if instead minimizing the output of $D$, which indicates the areas that largely contribute to the decision are the same for a discriminator, no matter positively or negatively.
The region with higher response within the attention map contributes more to the decision.
%

\begin{figure}[h]
\begin{center}
    \includegraphics[width=\linewidth]{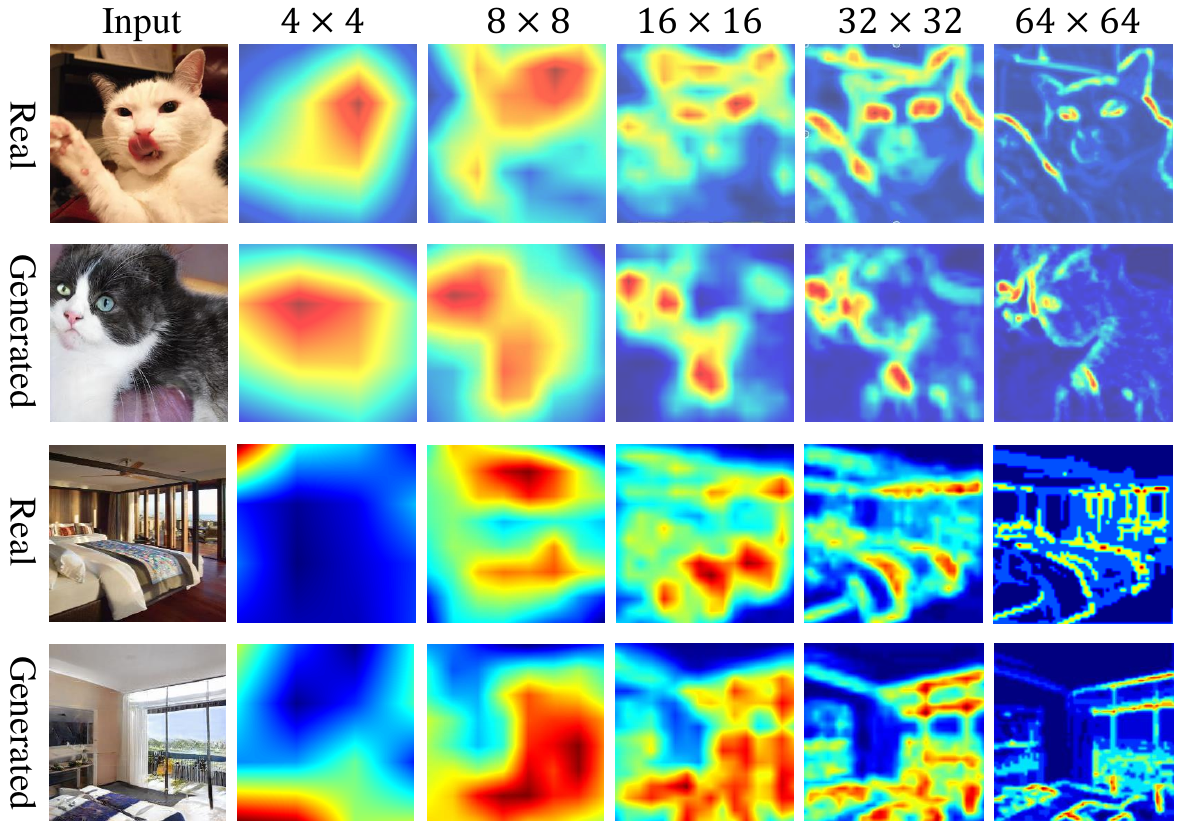}
    \caption{\textbf{Spatial visual attention at the intermediate layers of the discriminator}, visualized by GradCAM. A bright color indicates a strong contribution to the final score. `$64\times64$' indicates being upsampled from a  $64\times64$ feature map. The samples are the real images and the images generated by StyleGAN2~\cite{stylegan2}. 
    }    
    \label{Fig:Hie}
\end{center}
\vspace{-15pt}
\end{figure}

\begin{figure*}[htbp]
    \centering
    \includegraphics[width=\linewidth]{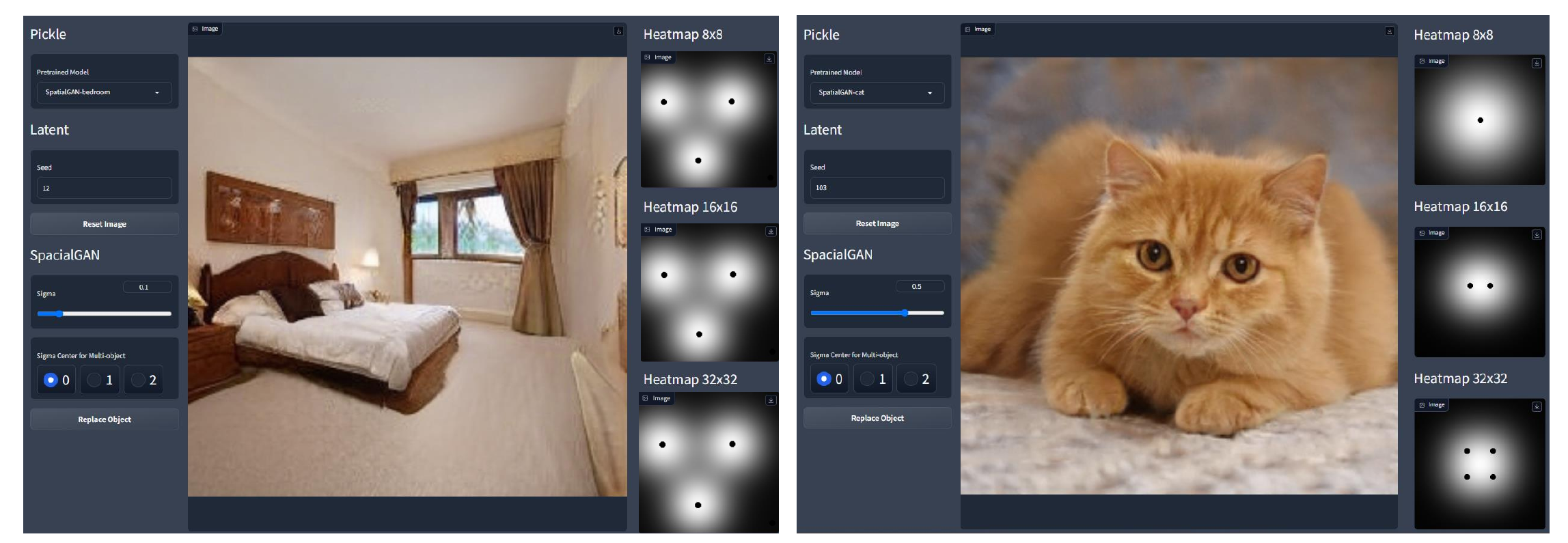}
    \caption{\textbf{User interface for interactive editing.}  Users can drag the Gaussian centers to alter heatmaps and synthesize an image corresponding to the new heatmap. In addition to that users can also change the sigma value of the heatmap center and the local style corresponding to that center. The Bedroom model (left) uses the same heatmap for all the resolutions while the Cat model (right) uses hierarchical heatmaps. \textit{Readers are suggested to view the demo video of interactive editing on our project page. }
    }
    \label{Fig:UI}
    \vspace{-5pt}
\end{figure*}
\cref{Fig:Hie} visualizes some GradCAM results under multiple feature resolutions.
Our examination of discriminators trained on datasets such as LSUN Cat and LSUN Bedroom reveals insightful aspects of their spatial behavior. 
We have following observations:
(1) $D$ learns its own visual attention on both real and generated images.
It suggests that $D$ makes the real/fake decision by paying more attention to some particular regions.
(2) The visual attention emerging from $D$ shows a hierarchical property.
In the shallow layers (like $64\times64$ and $32\times32$ resolutions), $D$ is attentive to local structures such as edge lines in the image.
As the layer goes deeper, $D$ progressively concentrates on the overall location of discriminative contents, \textit{e.g.}, the face of a cat.
(3) The hierarchical attention maps have fewer `local peaks' at more abstract feature layers with a lower resolution.
For example, there is only one peak in the $4\times4$ attention maps.


Building on this understanding, the discriminator of GANs has its own visual attention when determining a real or fake image. 
However, when learning to transform a latent vector into a realistic image, the generator receives no explicit clue about which regions to focus on.
Specifically, for a particular synthesis, $G$ has to decode all the needed information from the input latent code.
Furthermore, $G$ has no idea about the spatial preference of $D$ on making the real/fake decisions.
%
%
In~\cite{Wang_2022_CVPR}, we introduced the concept of deploying pseudo attention in G to enhance its synthesis capability and spatial steerability. In subsequent sections, we further refine our methods for both indoor and non-indoor scenes, leading to improvements in synthesis quality and spatial steerability.\\
%
%

\subsection{Encoding Spatial Heatmap in Generator}\label{subsec:spatial-awareness-to-g}

\vspace{2pt}
\noindent \textbf{Hierarchical Heatmap Sampling.}
Inspired by the observation in \cref{subsec:datt}, we propose a hierarchical heatmap sampling algorithm for single-object scenes.
The heatmap is responsible for teaching $G$ which regions to pay more attention to. 
Each heatmap is abstracted as a combination of several sub-regions and a background.
%
%
We formulate each sub-region as a 2D map, $H_i$, which is sampled subject to a Gaussian distribution,
\begin{equation}
    H_i \sim \mathcal{N} (\mathbf{c_i}, \mathbf{cov}), \label{eq:heatmap-sampling}
\end{equation}
where $\mathbf{c_i}$ and $\mathbf{cov}$ denote the mean and the covariance.
According to the definition of 2D Gaussian distribution, $\mathbf{c_i}$ just represents the coordinates of the region center.
The final heatmap can be written as the sum of all sub-maps, $H=\sum_{i=1}^{n}H_i$, where $n$ denotes the total number of local regions for $G$ to focus on.

As pointed out in the prior works~\cite{stylegan, higan}, the generator in GANs learns image synthesis in a coarse-to-fine manner, where the early layers provide a rough template and the latter layers refine the details.
To match such a mechanism and be consistent with the hierarchical spatial attention of discriminators, we design a hierarchical heatmap sampling algorithm.
Concretely, we first sample a spatial heatmap with \cref{eq:heatmap-sampling} for the most abstract level (\textit{i.e.}, with the lowest resolution), and derive the heatmaps for other resolutions based on the initial one.
The number of centers, $n$, and the covariance, $\mathbf{cov}$, adapt accordingly to the feature resolution. \\

\begin{figure}[htbp]
\vspace{-5pt}
\includegraphics[width=\linewidth]{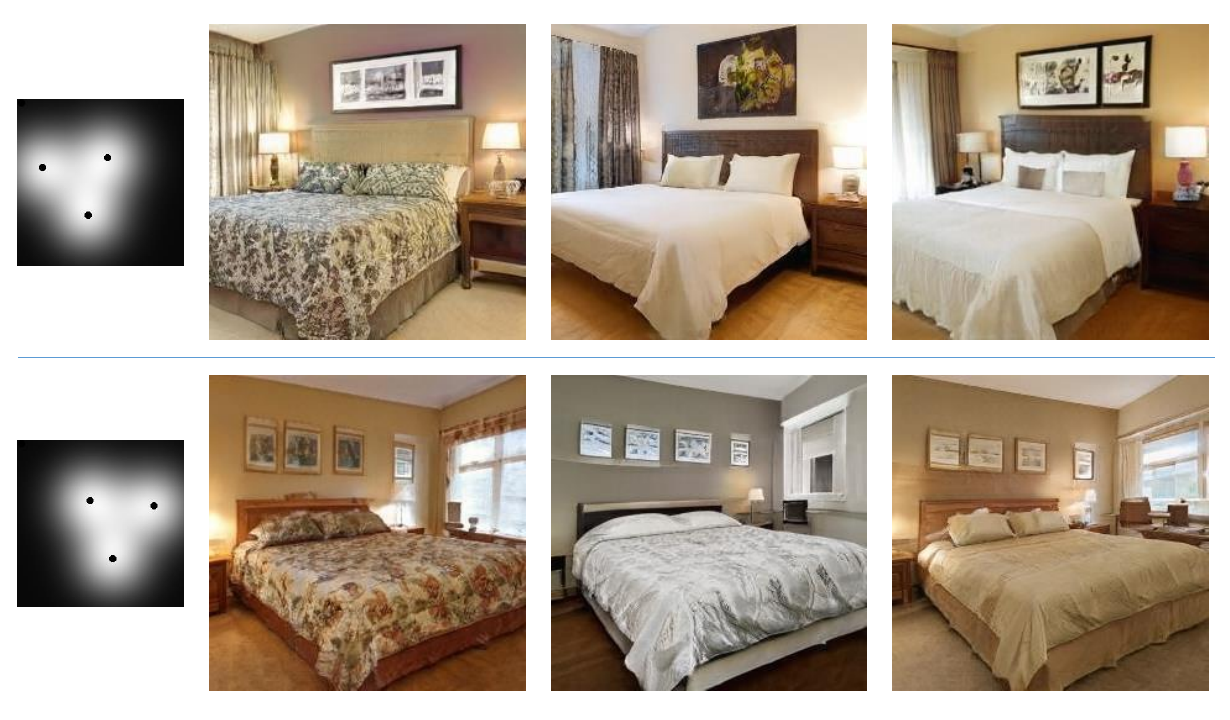}
    \caption{\textbf{Controlling the room layout by using the same spatial heatmap.} For each row, we adopt different latent codes but use the same spatial heatmap. The images in the same row show a similar spatial layout, while their appearances are different. The appearances include colour, texture, lighting, and so on. 
    }
    \label{Fig:Bed_layout}
\end{figure}

\noindent \textbf{Multi-object Heatmap Sampling.} Given that indoor scenes often have several independent objects, a straightforward idea is to model each object by a hierarchical set of Gaussian heatmaps as discussed above and encode such a complicated inductive bias into each layer of $G$.
However, with hierarchical sampling, modeling multiple objects in an image would lead to too many sub-regions at high-resolution, which would interact and possibly conflict with each other.
In practice, we find such a conflict would let the model confused and hinder the optimization during the training process. 
Especially, a small modification of a heatmap sub-region would spread to other sub-regions and result in a dramatic change over the output synthesis, which troubles the spatial editing.
%
Therefore, we abandon the hierarchical sampling for the multi-object setting. 
Instead, we model multiple objects by different Gaussian distributions at the coarsest resolution, and use this heatmap as the inductive bias for \textit{all} the synthesis layers of $G$, which can be noticed in our illustration for the user interface \cref{Fig:UI}.
The heatmap in fact points out the rough location of various objects, which can be viewed as an abstracted scene layout.
In this way, we can control the input heatmap to keep the scene layout, as shown in \cref{Fig:Bed_layout}. \\

\begin{figure}[b]
\begin{center}
\includegraphics[width=\linewidth]{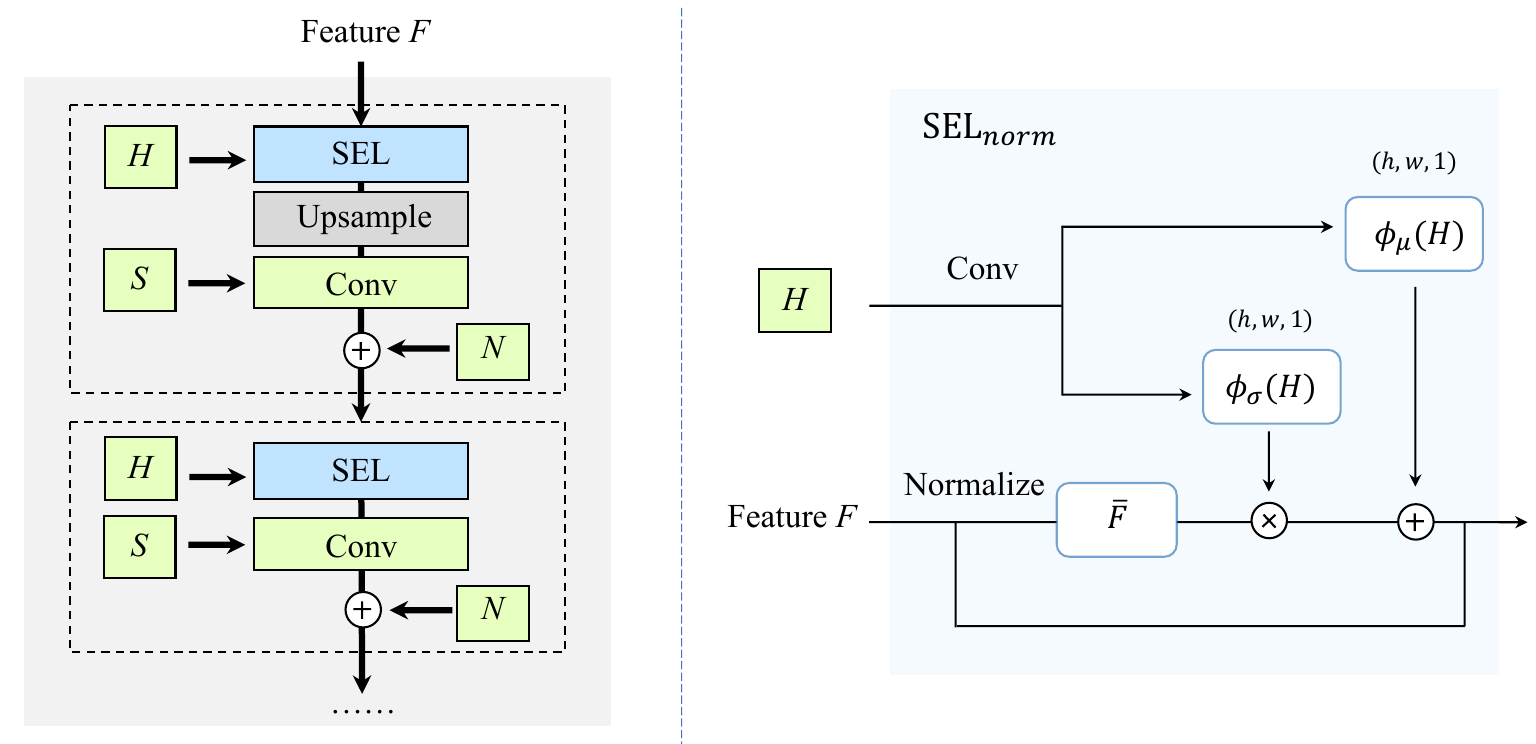}
  \caption{\textbf{Spatial Encoding,} where the left shows how the spatial encoding layer (SEL) works over StyleGAN2 at each resolution, and the right describes the internal of the SEL$_{\textit{norm}}$. The symbol `S' represents the style in StyleGAN2, `N' is the noise, and `H' indicates the spatial heatmap. Derived from ~\cite{spade}, we incorporate the spatial heatmaps into $G$ via normalization and denormalization. 
  }
\label{Fig:SEL}
\end{center}
\end{figure}

\vspace{2pt}
\noindent \textbf{Heatmap Encoding.}
We incorporate the spatial heatmaps into the intermediate features of $G$ to raise its spatial controllability (no matter for single-object or multi-object scenes).
It usually can be conducted in two ways, via feature concatenation or feature normalization~\cite{adain,spade}.
We use a spatial encoding layer (SEL), whose two variants are denoted as SEL$_{\textit{concat}}$ and SEL$_{\textit{norm}}$.
Specifically, the variant SEL$_{\textit{concat}}$ processes the concatenation of heatmap and feature via a convolution layer, and outputs new feature for the next layer.
Inspired by SPADE~\cite{spade}, the variant SEL$_{\textit{norm}}$ integrates the hierarchical heatmaps into the per-layer feature maps of $G$ with normalization and denormalization operations, as
\begin{equation}
    SEL_{\textit{norm}}(F,H) = \phi_{\sigma}(H) \  \frac{F- \mu(F)}{\sigma(F)} + \phi_{\mu}(H),
\end{equation}
where $F$ denotes an intermediate feature map produced by $G$, which is with the same resolution as $H$.
$\mu(\cdot)$ and $\sigma(\cdot)$ respectively stand for the functions of computing channel-wise mean and standard deviation.
$\phi_{\mu}(\cdot)$ and $\phi_{\sigma}(\cdot)$ are two learnable functions, whose outputs are point-wise and with a shape of $(h, w, 1)$.
%
%
Besides, as shown in \cref{Fig:SEL}, we use a residual connection to stabilize the intermediate features.
If not particularly specified, we adopt the variant SEL$_{\textit{norm}}$ since it shows a slightly better performance.  
%

It is worth noting, although we learn the SEL$_{\textit{norm}}$ architecture from SPADE~\cite{spade}, these two methods are clearly different since SPADE targets at synthesizing images based on a given semantic segmentation mask, whose training requires paired ground-truth data, while our model is trained with completely unlabeled data. Meanwhile, SEL$_{\textit{norm}}$ is just a replaceable component of our approach. \\
%
%

\vspace{2pt}
\noindent \textbf{Coarse Heatmap Processing.} 
To better understand the role of heatmap encoding, we study the mapping from the input heatmap to the synthesized image.
In our conference version~\cite{Wang_2022_CVPR}, the heatmap $H$ is sampled as the same size as the target image, processed by function $\phi_{\mu}(\cdot)$ and $\phi_{\sigma}(\cdot)$, and then down-sampled to match the resolution of intermediate features. 
It was designed to keep the heatmap continuous and consistent with the final output.
However, here we find it is not a superior choice.
Instead, if we first down-sample the fine-grained heatmap to the intermediate resolution and then process it via a convolution layer, the synthesis quality would be improved and the computation cost is decreased.
In other words, we process a `coarse' heatmap.
We speculate two reasons for this observation:  
(1) A spatially continuous inductive bias may not be necessary. The heatmap (sampled by a 2D Gaussian distribution) is continuous while the real attention could be discrete, because the real-world distribution of objects may be discrete.  %
The network possibly only needs a rough spatial structure instead of a fine-grained guidance.
%
(2) For heatmap processing, a global understanding matters. 
With the same convolutional layer, down-sampling the input in fact enlarges the receptive field by times, which enjoys a global view.
%
%
We verify these two hypotheses and provide detailed analysis in \cref{subsec:ablation}. \\







\noindent \textbf{Multi-object Heatmap Encoding.} 
%
%
%
\begin{figure}[b]
\begin{center}
\includegraphics[width=0.7\linewidth]{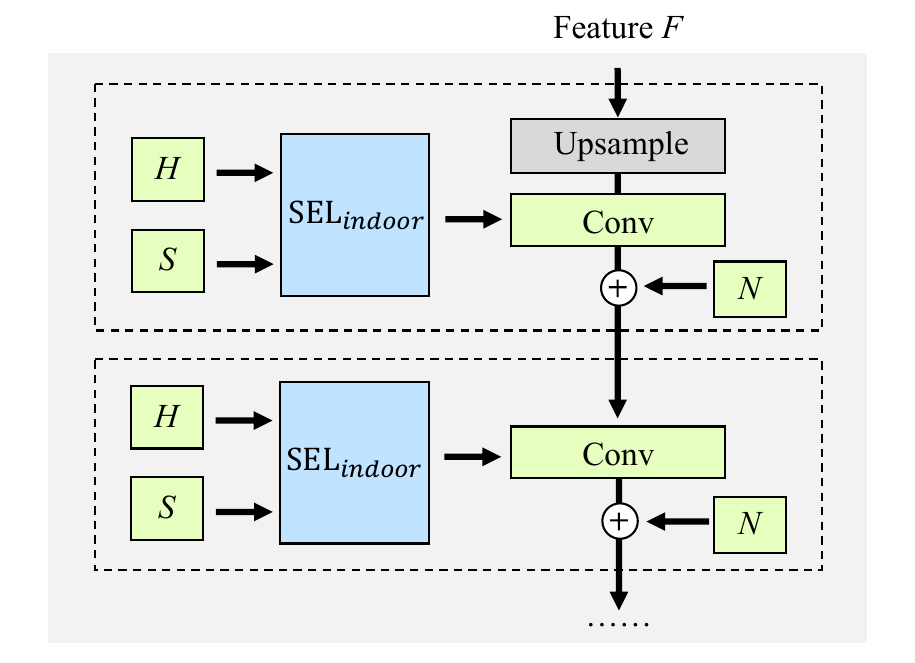}
  \caption{\textbf{Illustration of SEL$_{\textit{indoor}}$}, where we integrate the heatmap $H$ and style $S$ to avoid modifying the intermediate features twice like in \cref{Fig:SEL}.
  }
\label{Fig:SEL_indoor}
\end{center}
\end{figure}
%
%
To further improve the spatial steerability of the model, we review the heatmap encoding process.
It is worth noting that the intermediate features of $G$ are successively modified by heatmap $H$ and style $S$, as illustrated by the left of \cref{Fig:SEL}.
This architecture was designed following the philosophy that for an image, the spatial structure should be determined first and then comes to other components like appearance.
However, the feature change brought by styles may weaken the effect of spatial heatmaps.
We empirically find this phenomenon is obvious in indoor scenes, where we suppose the multi-object setting increases the difficulty of utilizing the encoded spatial heatmaps.
Therefore, we propose a variant SEL$_{\textit{indoor}}$, as shown in \cref{Fig:SEL_indoor}.
It integrates the spatial heatmaps $H$ and styles $S$ together and encodes their combination to $G$, which only modifies the generator features once. 
Since $H$ is a heatmap within the range $(0,1)$, the integration can be effectively conducted via an element-wise product, that 
%
    $\text{SEL}_{\textit{indoor}}(S,H) = S * H$,
%
from $(1, 1, c)$ and $(h, w, 1)$ to a shape of $(h, w, c)$. 
%

Moreover, since now one 2D Gaussian (sub-heatmap $H_i$) represents a separate object, we could employ a unique appearance code for each sub-heatmap, to reduce their interaction. 
We utilize $n$ style codes $S_i$ to depict the indoor objects and one $S_{bg}$ to control the background appearance:
 \begin{equation}
    SEL_{\textit{indoor}}(S,H) = \sum_{i=1}^{n}S_i * H_i + S_{bg}.
 \end{equation}
The value of $H_{i}$ would decay to zero in its edge areas and hence the corresponding appearance code $S_i$ would gradually lose the effect.
All the style codes $S_i$ and $S_{bg}$ share the same feature dimension.
For simplicity, we encode SEL$_{\textit{indoor}}(S,H)$ into $G$ as same as encoding $S$, just except being spatial-aware.
%
This new heatmap encoding method enhances the spatial steerability of the model, supporting the editing operations like removing an object or changing the style of a region, as shown in \cref{Fig:Manipulation}.
\\
\vspace{-5pt}
\subsection{Spatial Alignment via Self Supervision}\label{subsec:feedback-mechanism}

Encoding heatmaps into $G$ can explicitly raise its spatial steerability, but the heatmaps fed into $G$ are completely arbitrary.
%
%
Without further guidance, how $G$ is supposed to utilize the heatmaps is ambiguous, which influences the spatial steerability of the model.
For example, $G$ has no idea about `whether to pay \textit{more} or \textit{less} attention on the highlighted regions in the heatmap'.
Instead, $D$ learns its own visual attention based on the semantically meaningful image contents.
To make the best usage of the introduced spatial inductive bias, we propose to involve the spatial attention of $D$ as a self-supervision signal, which does not require any extra annotation.
Specifically, at each optimization step of $G$, we use $D$ to process the synthesized image and generate a corresponding visual attention map with the help of GradCAM.
Besides competing with $D$, $G$ is further trained to minimize the distance between the attention map induced from $D$ and the input heatmap $H$.
The loss function can be written as
\begin{align}  
\mathcal{L}_{\textit{align}} = || \  \text{GradCAM}_{D}[G(H, \mathbf{z})] - \ H \ ||_1.
\end{align} 
We truncate the $\mathcal{L}_{\textit{align}} $ values if smaller than a constant $\tau$, since the sampled heatmaps are not expected to perfectly match the real attention maps shaped by semantics.
The threshold $\tau$ is set as $0.25$ for all the experiments.
Note that $D$ is not updated in the process above and only used as a self supervision signal to train $G$.
Such a regularization loss aligns the spatial awareness of $G$ with the spatial attention of $D$, narrowing the information gap between them.
It employs $D$ to tell $G$ how to leverage the encoded inductive bias and hence raises the spatial controllability of $G$.
As an adversary, $D$ can also be a good teacher. \\
%


\vspace{2pt}
\noindent \textbf{Multi-object Self-supervision Objective.}
For conciseness, the sub-heatmaps for different indoor objects are treated equally in the sampling stage.
However, some objects may take over much attention in real-world situations, \textit{e.g.}, a very large bed.
Consequently, we relax the training objective for spatial awareness alignment.
Still leveraging discriminator attention as the self-supervision signal, we compare it to the heatmap $H$ and each sub-heatmap $H_i$, with the minimum distance as the optimization term:
\begin{align}  
\mathcal{L}_{\textit{align\_indoor}} = \min_{n+1}|| \  \text{GradCAM}_{D}[G(H, \mathbf{z})] - \ \widehat{H} \ ||_1,
\end{align} 
where $\widehat{H} \in \{H_{1}, H_{2}, ..., H_{n}, H\}$.
%
%
The value truncation and discriminator freezing in $\mathcal{L}_{\textit{align}}$ are still used here.
\subsection{Synergy between DragGAN and SpatialGAN}\label{subsec:DragGAN-x-sGAN}

In this section, we detail the integration of our SpatialGAN method with the latest point-based image manipulation technique, DragGAN \cite{pan2023drag}. DragGAN demonstrates proficiency in relocating specific parts of an image from one point to another.
%
Specifically, DragGAN assumes the feature space of a GAN model is discriminative enough to support precise point tracking and motion supervision, which is consistent with the observation in our paper. 
%
Therefore, the method optimizes the GAN latent code \( \mathbf{w} \) to encourage some  given handle points (\textit{i.e.}, starting points) to move towards their target destinations.
%
To ensure stable movement and precise manipulation, this process is typically repeated $30-200$ times.  



%
%

Direct application of DragGAN's techniques to SpatialGAN yielded inferior manipulation results, mainly because of the architectural differences between SpatialGAN and StyleGAN2, especially the inclusion of heatmaps in our design. 
We discovered that effective manipulation in SpatialGAN requires point-based optimization of both the latent space and the heatmaps. 
%
We have explored two distinct approaches for heatmap optimization: one by directly optimizing the pixels of  heatmaps $H$, and the other by focusing on the optimization of the heatmap centers $\mathbf{c}$. 
Preliminary results indicate that optimizing the centers of heatmaps yields more effective manipulation. 
%
By simultaneously optimizing heatmap centers and latent codes, we  not only augment DragGAN's capabilities to support SpatialGAN but also drastically reduces the optimization steps for image manipulation, requiring just $15-70$ steps
%
%


However, although DragGAN shows an impressive  ability for granular manipulation, its iterative optimization at inference is time-consuming. 
Unlike DragGAN, our SpatialGAN does not require any optimization process during inference. 
To leverage the strengths of both methodologies, we propose a two-step manipulation process: initially, our method is employed for coarse movement, followed by the application of DragGAN to refine the movement. 
Given the handle points and the target destinations, SpatialGAN will first manipulate the image by altering the heatmaps to enable the corresponding movement, which can skip a lot of iterative optimization steps required by DragGAN.  
Then, we adopt DragGAN over the manipulated image to ensure the handle points precisely match the target destinations.
%
Specifically, the coarse movement is guided by the condition: if the starting points are within a specified radius \( r \) of the heatmap centers \( c_i \), the centers are moved to new positions \( c'_i \) according to the formula:
\begin{equation}
 c'_i = c_i + \alpha \cdot (t - p) ,
\end{equation}
where \( p \) is the start point, \( t \) is the target point, and \( \alpha \) is a factor determining the extent of movement towards the target. In cases where the starting points are outside the radius, the heatmap centers remain unchanged.
%
Following this, point-based optimization takes over for fine manipulation, where the heatmap centers and latent code are jointly optimized. The process, described here for a single handle point, can be easily extended to multiple points. This hybrid methodology significantly reduces the optimization steps required for precise alignment, harnessing the strengths of both SpatialGAN's efficient coarse adjustment and DragGAN's fine manipulation. As a result, the optimization steps are further condensed to approximately $10-20$ iterations, greatly diminishing the time required for manipulation. 
%



\begin{figure*}[t]
    \centering
    \includegraphics[width=\linewidth]{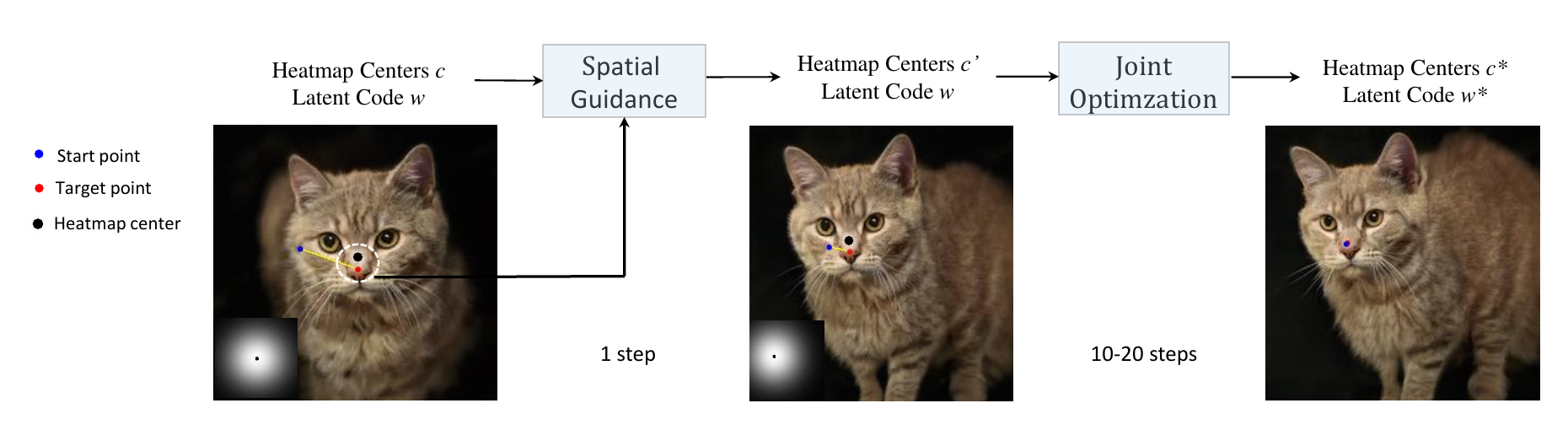}
    \vspace{-5pt}
    \caption{\textbf{Overview of SpatialGAN + DragGAN pipeline.} In this example, the cat's face is initially marked for repositioning. Subsequently, a coarse spatial adjustment is performed by shifting heatmap centers. Then we jointly optimize the heatmap centers and latent codes, to culminate in the precise alignment of features with predetermined targets, highlighting the model's fine-tuning abilities for detailed editing.
    }
    \label{Fig:Method_sgan_dragGAN}
\end{figure*}

The successful integration of SpatialGAN with DragGAN demonstrates the adaptability of our method, illustrating its compatibility with various manipulation techniques. This integration highlights the potential of SpatialGAN as an advanced toolkit for the community engaged in spatial editing within generative models. The comprehensive methodology of our approach, is systematically illustrated in Figure~\ref{Fig:Method_sgan_dragGAN}.

%

\section{Experiments}
\label{sec:exp}

We evaluate the proposed SpatialGAN on multiple benchmarks, covering faces, indoor scenes, and outdoor scenes. 
Sec.~\ref{subsec:impl} provides the implementation details. 
The main comparison and experimental results are presented in Sec.~\ref{subsec:main-results}, that our SpatialGAN can support multiple types of editing, enhance the spatial controllability of $G$, and improve the synthesis quality.
Sec.~\ref{subsec:ablation} includes comprehensive ablation studies.

\subsection{Implementation Details}\label{subsec:impl}

\noindent \textbf{Datasets.} We conduct the experiments on the FFHQ~\cite{stylegan}, LSUN Cat~\cite{yu2015lsun}, Church~\cite{yu2015lsun}, and Bedroom~\cite{yu2015lsun} datasets.
%
%
The FFHQ dataset consists of $70K$ high-resolution ($1024\times1024$) images of human faces, under Creative Commons BY-NC-SA 4.0 license~\cite{FFHQ_github}. Usually, the images are horizontally flipped to double the size of training samples. 
The LSUN Cat dataset contains $1600K$ real-world images regarding different cats, the LSUN Church dataset includes $126K$ images with church scenes, and the LSUN Bedroom dataset provides around $300K$ complex indoor bedroom images from different views.
%
%
Following the setting of~\cite{stylegan2-ada}, we respectively take $200K$ LSUN Cat samples and $200K$ LSUN Bedroom images for training.
We use all the FFHQ and LSUN Church images for training.
It is worth noting that all images are resized to $256 \times 256$ resolution. 
\\

\noindent \textbf{Spatial Heatmap Sampling and Encoding.}
In practice, we find the GradCAM maps on the fine resolutions are too sensitive to semantic cues.
Therefore, we only conduct encoding on the level $0,1,2$ of $G$, \emph{i.e.}, resolution $4\times4$, $8\times8$, and $16\times16$.
For non-indoor scenes, we heuristically generate $1,2,4$ centers (in other words, sub-heatmaps) on these three levels.
The center for the level $0$ heatmap, denoted as $\mathbf{c^{0}_0}$, is sampled using a Gaussian distribution with a mean positioned at half the height and width $(\frac{h}{2},\frac{w}{2})$, and a standard deviation of a third of the height and width $(\frac{h}{3},\frac{w}{3})$.
To maintain heatmap consistency across different levels, the centers for levels $1$ and $2$ ($\mathbf{c^{1}_k}$ and $\mathbf{c^{2}_k}$) are sampled based on the position of the level $0$ center.
%
%
In this context, $\mathbf{c^{1}_k}$ and $\mathbf{c^{2}_k}$ represent multiple potential centers at levels $1$ and $2$, respectively, with `$k$' indicating the specific center number.
The standard deviations for these are set at $(\frac{h}{6},\frac{w}{6})$.
It's important to note that if the center at level $0$ shifts, the heatmaps at other levels will adjust accordingly.
Following the coarse-to-fine manner, we decrease each center's influence area level by level.
Besides, we drop the sampling if the level $0$ center is outside the image.
In our observation, the results of the proposed method are robust to these hyperparameters for heatmap sampling.
Therefore, we use the same hyperparameters for heatmap sampling on the FFHQ, LSUN Cat, and LSUN Church datasets. 
For the indoor setting, we take $n=3$ sub-heatmaps to construct a complete heatmap, whose centers are sampled via a Gaussian distribution with a mean of $(\frac{h}{2},\frac{w}{2})$ and a standard deviation of $(\frac{h}{2},\frac{w}{2})$. 
We encode this heatmap to all the synthesis layers of $G$.
More implementation details are provided in the Supplementary Material.   \\

\noindent \textbf{Training and Evaluation.} We implement our SpatialGAN on the official implementation of \href{https://github.com/NVlabs/stylegan2-ada-pytorch}{StyleGAN2}, such that the state-of-the-art image generation method StyleGAN2~\cite{stylegan2} serves as our baseline.
We follow the default training configuration of \cite{stylegan2-ada} for the convenience of reproducibility, and keep the hyperparameters unchanged to validate the effectiveness of our proposed framework.
For example, we train all the models with a batch size of $64$ on $8$ GPUs and continue the training until $25M$ images have been shown to the discriminator. 
Our method increases the training time by around $30\%$ compared with the baseline. 
We use Fréchet Inception Distance (FID)~\cite{heusel2017gans} between $50K$ generated samples and all the available real samples as the image generation quality indicator.
\\

%
%
%
%
%
%
%
%
%
%
%
%
%
%
%
\subsection{Main Results}\label{subsec:main-results}
\begin{figure*} 
    \centering
    \includegraphics[width=\linewidth]{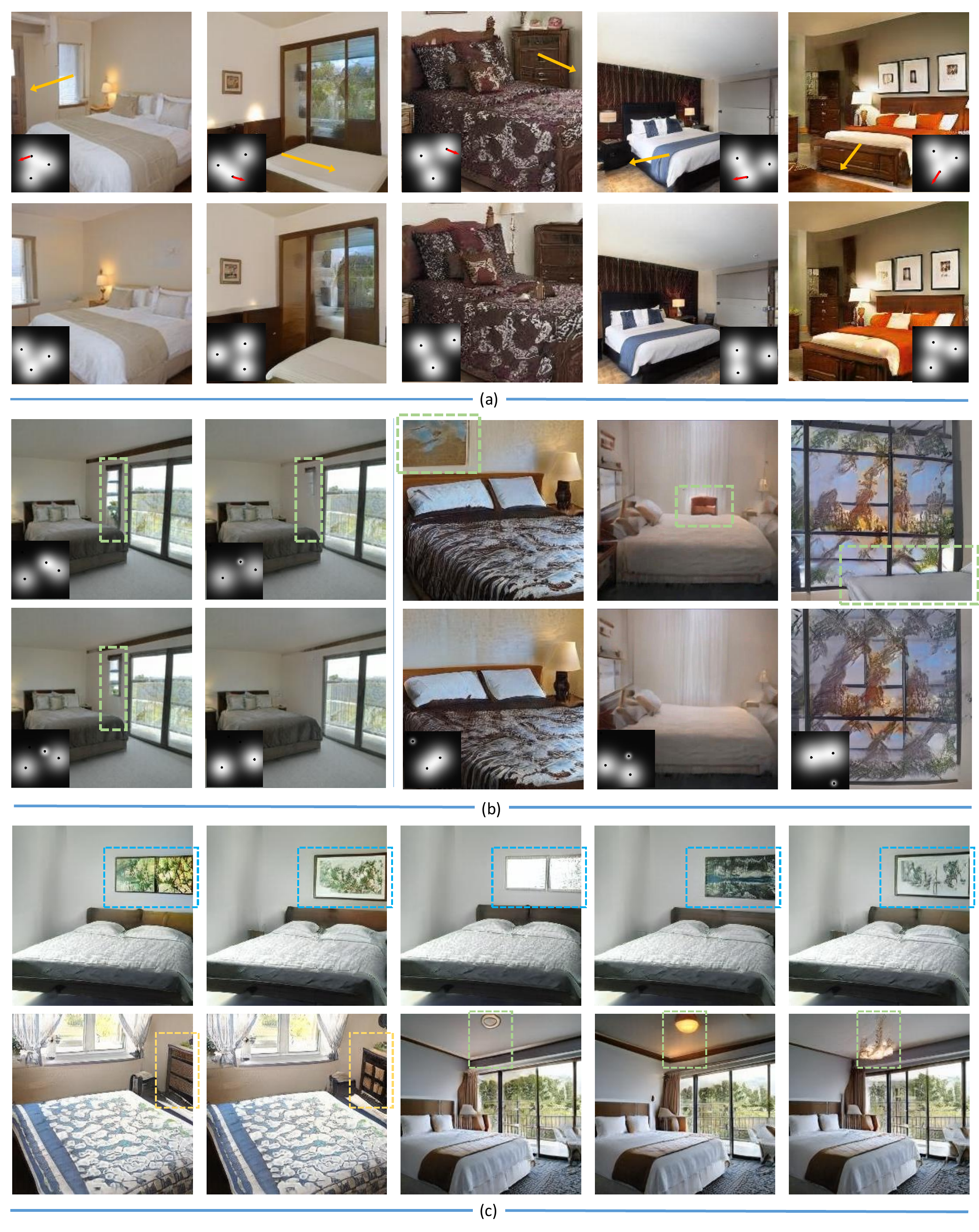}
    \caption{\textbf{Manipulating Multi-Object Indoor Scenes.} \textit{(a)} \textbf{Rearranging Objects:} By moving one sub-heatmap center (yellow arrows), corresponding objects like windows and beds are disentangled and moved, with the generator adjusting nearby regions for coherence. \textit{(b)} \textbf{Removing Objects:} Objects can be removed by eliminating their associated sub-heatmaps, as shown by the gradual removal of elements like windows and beds, leaving the background and other objects mostly unchanged. \textit{(c)} \textbf{Replacing Objects:} The appearance of local regions is altered using unique style codes for each sub-heatmap, enabling variations in paintings, windows, and light types, as indicated by the blue and green boxes.}
    \label{Fig:Bed_combine}
\end{figure*}


\noindent \textbf{Enhanced Spatial Controllability}. 
Building on our previous discussion, we introduced spatial awareness into the generator ($G$), aiming to enhance its spatial steerability. 
In this section, we present qualitative results from various datasets to demonstrate that $G$ effectively concentrates on areas indicated by the input heatmaps, thereby facilitating a range of spatial editing applications.
For the indoor scenes, by keeping the spatial heatmaps unchanged, we could control the overall layout (\cref{Fig:Bed_layout}). The bedrooms generated with the same heatmap (each row) will arrange objects in a similar manner, even though the object semantics vary.
Furthermore, for the indoor scenes with multi-object heatmap sampling, we can move objects as shown in \cref{Fig:Bed_combine} (a). For example, in the right most column of \cref{Fig:Bed_combine} (a), we drag the sub-heatmap center to the left, and the bed correspondingly moves to the left, where the object trajectory is denoted by a yellow arrow. It is also worth noting that $G$ could adaptively modify the nearby texture and structure to produce a reasonable image.

Moreover, since we model the objects with independent sub-heatmaps and style codes for the indoor setting, it allows removing objects or changing the style of a partial region.
As shown in \cref{Fig:Bed_combine} (b), we can remove various objects in the indoor scenes.
For example, in the left two columns, a window is gradually removed as we decrease the area of the associated sub-heatmap.
In other columns, the objects like a painting or a bed are removed.
%
Although the model slightly adjusts the nearby region to keep the synthesis reasonable, the overall image is unchanged. 
The partial editing samples are shown in %
\cref{Fig:Bed_combine} (c).
As mentioned in \cref{subsec:spatial-awareness-to-g}, we can edit the style code of a specific sub-heatmap and hence change the appearance of the corresponding region.
For instance, we can edit the region specified by the blue boxes to different types of paintings and windows.
%

\begin{figure*}[htbp]
    \centering
    \includegraphics[width=\linewidth]{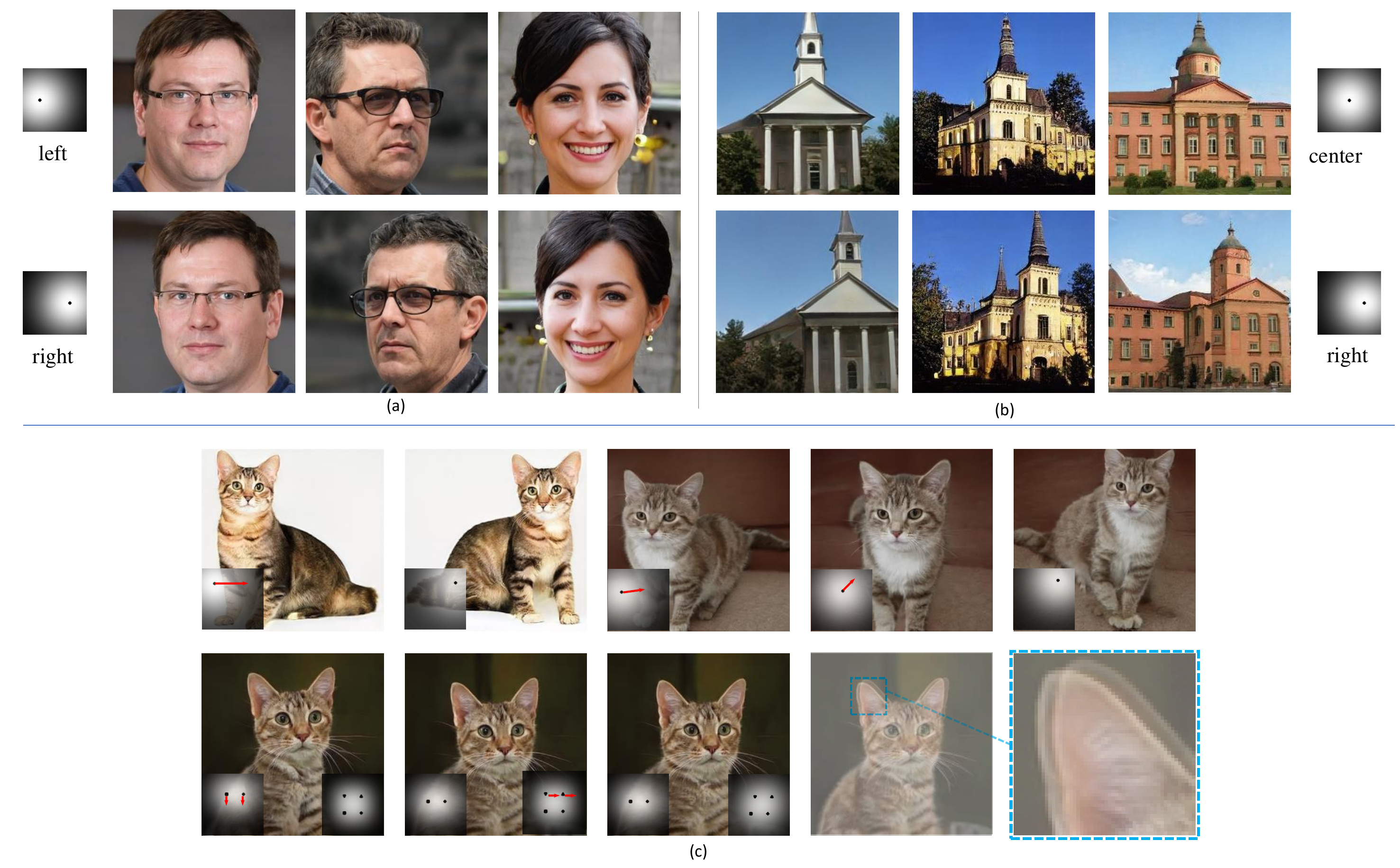}
    \caption{\textbf{Qualitative results on the FFHQ (a), the LSUN Church (b), and LSUN Cat dataset (c)}. Each row uses the same spatial heatmap but different latent codes, and each column uses the same latent code. We can see that the spatial heatmap roughly controls the pose of the face and the viewpoint of the church building, which facilitates the interactive spatial editing of the output image. To further show the hierarchical structure, we move the heatmap to the fine-grained level in the last row. Different from the body movement, the change in $8\times8$ heatmap (two centers) mainly moves the cat eyes, and the change in $16\times16$ heatmap (four centers) leads to subtle movement of the cat ears. It is worth noting that, as the content is being manipulated, our $G$ knows to adjust the nearby regions to make everything coherent.  
    } 
    \label{Fig:Single_obj_combine}
\end{figure*}

Turning our attention to single object scenes, we find that by keeping the spatial heatmaps unchanged, we can control the pose of human faces and the viewpoint of churches \cref{Fig:Single_obj_combine} (a and b). For instance, as we move the level $0$ heatmap of the sample in row 1 in ~\cref{Fig:Single_obj_combine} (c), the cat bodies move under the guidance of heatmap movement, indicated by red arrows.
%
In the second row of ~\cref{Fig:Single_obj_combine} (c), the change in level $1$ heatmap leads to a movement in cat eyes. As we slightly push the top two centers of the level $2$ heatmap to the right, the cat ears subtly turn right while other parts, even the cat whiskers, remain unchanged. Overall, these qualitative samples, spanning both indoor and non-indoor scenes, demonstrate the spatial steerability introduced by our SpatialGAN.\\

\noindent \textbf{Quantitative Analysis of Spatial Steerability.} 
In order to assess our method's spatial steerability in a quantitative manner, we examine how the movement of objects is influenced by alterations in the heatmap. 
Ideally, if the center of a sub-heatmap $H_i$ moves by certain pixels, the corresponding object should also move by the same pixels in the same direction.
With the help of the off-the-shelf instance segmentation model like Mask RCNN~\cite{he2017mask}, we can roughly quantify the movement of a specific object.
In detail, for one sub-heatmap $H_i$, we view its corresponding object as the one where $H_i$ center lies in.
Its object center is the average position of the pixels belonging to this object, segmented by Mask RCNN.
We move the $H_i$ center by $\mathbf{p}$ and generate a new image, where $\mathbf{p}$ is a random 2D vector.
Assuming the overall appearance remains unchanged, we traverse the objects in the new image and look for the object with the smallest feature distance to the one in the original image.
We view it as the moved object and compute the vector of its center movement as $\mathbf{q}$.
Project $\mathbf{q}$ into the $\mathbf{p}$ to get its movement in the desired direction, $\frac{\mathbf{p} \cdot \mathbf{q}}{ \| \mathbf{p}  \| } $ as a scalar.
We pick its ratio over the desired movement scalar, $\frac{\mathbf{p} \cdot \mathbf{q}}{ {\| \mathbf{p}  \|}^2 } $, as the indicator of how effective the heatmap controls the synthesis, named as Co-move Ratio.
The closer this indicator is to 1, the better.
We evaluate a model's Co-move Ratio by averaging the results of $50k$ synthesis samples.
The ablation study was conducted on the indoor dataset LSUN Bedroom because it contains multiple independent objects and there are publicly available Mask RCNN models trained on similar scenes.

The results are provided in \cref{tab:comove_ratio}. 
The baseline StyleGAN2 is denoted as `N/A' because it does not support such a moving manipulation.
It is noticed that our conference version can move objects but just roughly, with a Co-move Ratio of $0.32$, because this version was not designed for scenes with multiple independent objects.
By adopting multi-object heatmap sampling, we mitigate the sub-heatmap conflict mentioned in \cref{subsec:spatial-awareness-to-g} and increase the ratio to $0.41$.
The largest improvement, from $0.41$ to $0.62$, is brought by SEL$_{indoor}$, where we (1) integrate the spatial heatmaps $H$ and styles $S$ to ensure only modifying the intermediate features once, and (2) employ a unique appearance code for each sub-heatmap.
The indoor self-supervision objective further improves the ratio to $0.69$.
These results verify the effectiveness of our designs for multi-object scenes, which improves the model's ability to move objects as desired. \\

\begin{table}[t]
\small
\begin{center}
\caption{\small \textbf{Quantitative analysis on the improvement of spatial controllability.} The experiments were conducted on the LSUN Bedroom dataset. Co-move Ratio indicates the ability to spatially move objects, the closer to 1 the better.  }
\setlength{\tabcolsep}{12pt}
\begin{tabular}{c|c}
\hline
 Method  & Co-move Ratio \\
 \hline
 Baseline  & N/A \\
 Ours Conf~\cite{Wang_2022_CVPR} & 0.32 \\
 + Multi-obj Sampling & 0.41 \\
+ SEL$_{indoor}$ & 0.62 \\ 
+ $\mathcal{L}_{\textit{align\_indoor}}$ & 0.69 \\
\hline
\end{tabular}
\label{tab:comove_ratio}
\end{center}
\vspace{-10pt}
\end{table}


\begin{figure}[h]
    \centering
    \includegraphics[width=\linewidth]{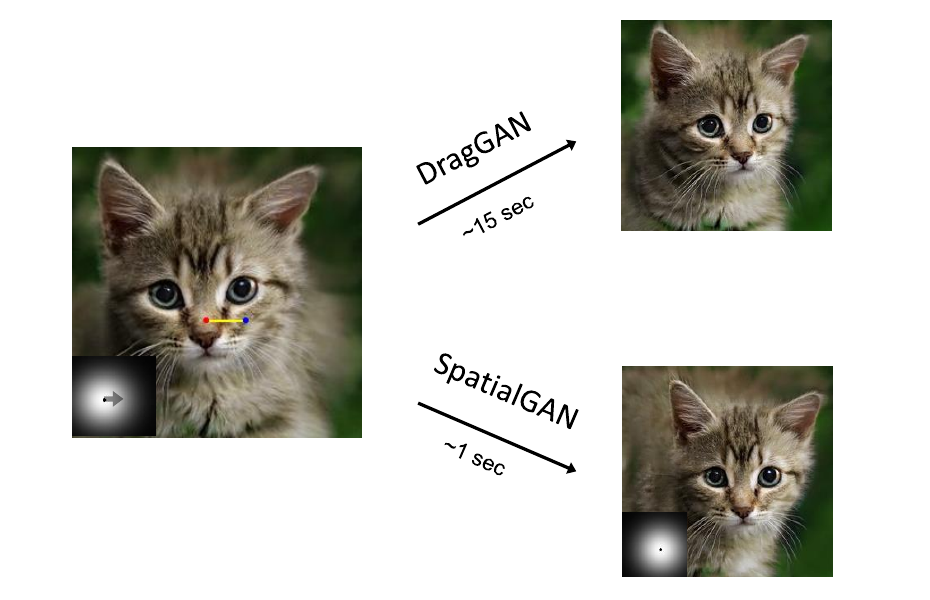}
    \caption{\textbf{Speed Comparison between DragGAN and SpatialGAN.} The images illustrate shifting a cat's face a few pixels to the right. While DragGAN requires approximately 15 seconds to complete this task, our SpatialGAN accomplishes the same in just 1 second.  
    }
    \label{Fig:DragGAN_x_sGAN_speed_comparision}
\end{figure}

\noindent \textbf{Comparison of Manipulation Ability.} As discussed before, we design a two-stage manipulation strategy to leverage the benefits of both SpatialGAN and DragGAN, which largely reduces the time required for point-based manipulation. 
%
Here we provide a quantitative analysis of the speed  in \cref{tab:speed}, averaging the results from $50$ manipulation samples. 
%
For example, for the LSUN Cat dataset, moving the cat's face by $60$ pixels takes DragGAN approximately $31$ seconds and $30-200$ optimization steps. 
Instead, the combination of SpatialGAN and DragGAN not only reduces the manipulation time to $11$ seconds  but also significantly decreases the optimization steps to $10-20$. 
%
%
This hybrid two-stage approach offers both coarse and fine manipulation capabilities, providing a more efficient, dynamic, and versatile tool for image manipulation tasks. Please also notice that, without point-based optimization, our SpatialGAN can conduct manipulation within just one second, although it does not support granular manipulation.


In addition to speed comparisons, we evaluated the effectiveness of our SpatialGAN-DragGAN hybrid approach in terms of the relative distance between the start and target points during image manipulation tasks. With DragGAN alone, the final average distance relative to the image size was approximately $0.015$. When employing our SpatialGAN combined with DragGAN, this relative distance averaged around $0.027$. This result demonstrates that, while our method is three times quicker than DragGAN alone, both methods exhibit comparable accuracy in spatial manipulation.



\begin{table}[h]
\small
\begin{center}
\caption{\textbf{Quantitative Analysis of Speed Performance.} This table presents the results of speed comparison experiments conducted using the LSUN Cat dataset. It showcases the average time required to shift the cat's face by a specified pixel distance. }
\begin{tabular}{c|c |c |c  }
\hline
\multirow{2}{*}{Method} & {Move} & {Move} & {Optimization}  \\
 & 30 pix & 60 pix & steps  \\
\hline
DragGAN ~\cite{pan2023drag} & 15 sec & 31 sec & 30-200    \\
Heatmap \& Latent Code & 18 sec & 35 sec & 15-70  \\
SpatialGAN + DragGAN  & \textbf{6 sec }& \textbf{11 sec} & \textbf{10-20}  \textbf{}\\
\hline
\end{tabular}
\label{tab:speed}
\end{center}
\vspace{-10pt}
\end{table}

\noindent \textbf{Interactive Interface}.
To further enhance user engagement and control, we have developed an interactive interface (UI), as illustrated in~\cref{Fig:UI}. Compared to the conference version, this interface is versatile, supporting models trained under both multi-object indoor and single-object settings. Moreover, utilizing a well-trained SpatialGAN model, users can initiate the process by selecting a random seed to generate an initial image. The interface automatically generates heatmaps with preset centers and sigma values. Users can interactively modify these heatmaps by clicking and dragging the centers, triggering real-time image synthesis reflective of these adjustments. The heatmap alterations are displayed in sequence, corresponding to feature resolutions of $4\times4$, $8\times8$, and $16\times16$, or levels $0$, $1$, and $2$, visible on the interface's right side. For the indoor model, a consistent multi-object heatmap is applied across all layers, while for other models, a hierarchical heatmap structure is employed. Another addition to our interface is the support to adjust the heatmap areas. In the context of the multi-object setting, this feature offers the ability to manipulate the size or even facilitate the removal of objects associated with specific heatmap centers. Furthermore, for nuanced control in the multi-object setting, users can select specific centers to alter their corresponding heatmap areas.

Additionally, our user interface now features integration with the DragGAN optimization pipeline, enhancing the granularity of user control. As discussed in \cref{subsec:DragGAN-x-sGAN}, the combined optimization process allows users to precisely adjust object placements with DragGAN, as well as to perform coarse manipulations with SpatialGAN. Consequently, the UI provides an intuitive platform for engaging with the advanced features of the system, streamlining the user experience in creative image manipulation.


\begin{table}[t]
\small
\begin{center}
\caption{\textbf{Quantitative results on the LSUN Cat, FFHQ, LSUN Church, and LSUN Bedroom datasets, all trained with $\mathbf{25}\mathbf{M}$ images shown to discriminator.} The baseline uses the architecture of StyleGAN2~\cite{stylegan}. We use FID as the metric for image generation quality.
$\downarrow$ denotes smaller is better. }
\setlength{\tabcolsep}{7pt}
\begin{tabular}{c|c |c |c | c}
\hline
\multirow{1.5}{*}{Method} & {Bedroom} & {Cat} & {FFHQ} & {Church} \\
& FID $\downarrow$ & FID $\downarrow$ & FID $\downarrow$ & FID $\downarrow$ \\
\hline
Baseline & 4.27 & 8.36 & 3.66 & 3.73 \\
EqGAN-SA~\cite{Wang_2022_CVPR} & 2.95 & 6.81 & 2.96 & 3.11 \\
SpatialGAN & \textbf{2.72} & \textbf{6.57} & \textbf{2.91} & \textbf{2.86} \\
\hline
\end{tabular}
\label{tab:main_results}
\end{center}
\vspace{-10pt}
\end{table}

\noindent \textbf{Enhanced Synthesis Quality}. 
SpatialGAN, our proposed model, notably elevates the quality of synthesis. We present quantitative evidence of this improvement in \cref{tab:main_results}. Relative to the  baseline StyleGAN2 and our conference version EqGAN-SA, SpatialGAN demonstrates consistent enhancements across a variety of datasets. 
%
%
%
For the indoor dataset LSUN Bedroom, SEL$_{indoor}$ and $\mathcal{L}_{\textit{align\_indoor}}$ show an impressive performance, improving the baseline FID from $4.27$ to $2.72$. 
It is also better than our conference version, whose FID is $2.95$, which shows the advantage of the specific design for multi-object scenes. 
%
%
Transitioning to non-indoor datasets, the LSUN Cat dataset saw an FID improvement from $8.36$ to $6.57$. For context, our conference version had an FID of $6.81$ for this dataset. Similarly, the Church dataset's FID was reduced from $3.73$ to $2.86$, surpassing the conference version's FID of $3.11$.
%
%
The results on the LSUN Cat, FFHQ, and LSUN Church datasets are slightly better than our conference version~\cite{Wang_2022_CVPR} because we adopt the coarse heatmap processing introduced in \cref{subsec:spatial-awareness-to-g}.
%
%
\\

\subsection{Analysis and Discussion}\label{subsec:ablation}



\begin{table}[htbp]
\small
\begin{center}
\caption{\textbf{Ablation study of heatmap sampling on the single-object scenes.} We separately throw random Gaussian noise, spatial heatmap without hierarchical sampling, and our hierarchical heatmap as the input to the spatial encoding layer. The hierarchical sampling consistently shows a better performance. }
\setlength{\tabcolsep}{8pt}
\begin{tabular}{c|c|c|c|c}
\hline
FID $\downarrow$  & Baseline & Gau. Noise & Non-Hie & Hie  \\
\hline
Cat & 8.36&8.33& 7.03 & \textbf{6.57}\\
FFHQ & 3.66& 3.67 & 3.29 & \textbf{2.91}\\
Church & 3.73 & 3.69  & 3.20  & \textbf{2.86} \\
%
\hline
\end{tabular}
\label{tab:ab_heatmap}
\end{center}
\vspace{-10pt}
\end{table}

\noindent \textbf{Hierarchical Heatmap Sampling for Single Object Scenes.} 
%
We conduct an ablation study to validate the effect of hierarchical heatmap sampling on non-indoor scenes, as provided in ~\cref{tab:ab_heatmap}. 
Specifically, 2D Gaussian noise is first considered as a straightforward baseline experiment since it provides non-structured spatial information. Accordingly, 2D Gaussian noise introduces no performance gains. It indicates, merely feeding a spatial heatmap but without any region to be emphasized is insufficient to raise spatial awareness.

Besides, we also use multiple-resolution spatial heatmaps but discard the hierarchical constraint, referred as Non-Hie in~\cref{tab:ab_heatmap}. Namely, spatial heatmaps at different resolutions are independently sampled. The baseline is obviously improved by this non-hierarchical spatial heatmap, demonstrating the effectiveness of the spatial awareness of $G$. Moreover, when our hierarchical sampling is adopted, we observe further improvements over the synthesis quality. \\

%
%

\noindent \textbf{Spatial Encoding for Single Object Scenes.} 
In order to raise the spatial controllability of $G$, there exist several alternatives to implement spatial encoding. Therefore, on non-indoor scenes, we conduct an ablation study to test various methods. 
For example, the first way of feeding the spatial heatmap is to flatten the 2D heatmap as a vector, and then concatenate it with the original latent code. This setting aims at validating whether maintaining 2D structure of spatial heatmap is necessary. Besides, we also use two different SEL modules (\emph{i.e.}, SEL$_{\textit{concat}}$ and SEL$_{\textit{norm}}$) mentioned in Sec.~\ref{subsec:spatial-awareness-to-g}. 
\cref{tab:ab_spatial} presents the results. 
Apparently, simply feeding the spatial heatmap but without the explicit $2D$ structure leads to no gains compared to the baseline. It might imply that it is challenging to use a vector (like the original latent code) to raise the spatial awareness of the generator. Instead, the proposed SEL module could introduce substantial improvements, demonstrating the effectiveness of the encoding implementation. For a fair comparison, all the ablation studies use $\mathcal{L}_{\textit{align}}$. \\

\begin{table}[htbp]
\small
\begin{center}
\caption{\small \textbf{Ablation study of spatial encoding on single-object scenes.} We flatten the spatial heatmap and incorporate the vectorized one into latent code, denoted as `Flatten'. It destroys the $2D$ space structure, and hence cannot improve over the baseline. Instead, encoding heatmaps in the spatial domain is beneficial. The two variants of SEL show a similar result, where SEL$_{\textit{norm}}$ is slightly better. }
\setlength{\tabcolsep}{12pt}
\begin{tabular}{c|c|c|c}
\hline
 \multirow{2}{*}{Method}  & {Cat} & {FFHQ} & Church\\
 & FID $\downarrow$  & FID $\downarrow$ & FID $\downarrow$  \\
\hline
Baseline & 8.36  &3.66 & 3.73 \\
Flatten  & 8.52 & 3.75  & 3.79 \\
SEL$_{\textit{concat}}$ & 6.73  & 3.02 & 2.93  \\
SEL$_{\textit{norm}}$  & \textbf{6.57} & \textbf{2.91} & \textbf{2.86} \\
\hline
\end{tabular}
\label{tab:ab_spatial}
\end{center}
\vspace{-10pt}
\end{table}

\begin{figure}[htbp]
    \centering
    \includegraphics[width=\linewidth]{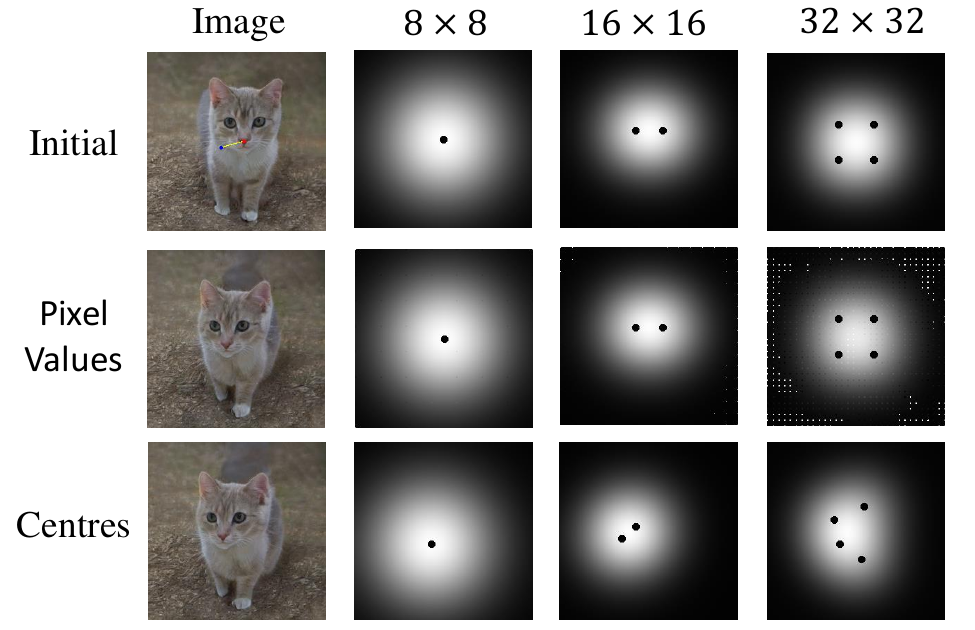}
    \caption{\textbf{Illustrative comparison of optimization strategies in DragGAN.}  \textbf{Initial Optimization (Top):} Shows the Initial image. \textbf{Heatmap Adjustment (Middle):} Demonstrates the effect of modifying the heatmap values, which leads to selective adversarial heatmap adjustments. \textbf{Center Optimization (Bottom):} Depicts the outcome of optimizing heatmap centers, which results in efficient and harmonious image manipulation. The columns represent the progression of heatmaps. }
    \label{Fig:DragGAN_heatmap}
\end{figure}

\noindent \textbf{Exploring Heatmap Dynamics in Point-based Optimization.} As previously discussed in the \cref{subsec:DragGAN-x-sGAN}, the integration of DragGAN into our method involves the simultaneous optimization of heatmap and latent code for point-based manipulations. In the beginning, we observed that optimizing only the latent code  — while keeping heatmap points static for single-object scenarios — imposes limitations. The unchanged heatmap restricts image transformation, while the evolving latent code attempts to induce change. This discrepancy hinders effective manipulation, leading to distorted outcomes without meaningful alteration.

Consequently, we experimented with modifying the pixel values of heatmaps $H$ together with the latent code. 
This approach resulted in selective pixel adjustments. As depicted in Figure \ref{Fig:DragGAN_heatmap}, we can observe that only certain pixel values of the heatmaps would be changed, and such changes mostly happen in the $32\times32$ level. 
%
Although this strategy achieved the desired image manipulation, its lack of explainability and the challenge it poses for further user manipulation are significant drawbacks.


To synergize with our spatial GAN method, we then optimize the heatmap centers, guiding them towards the intended directional movement of images. This strategy enables the heatmap to not only accommodate but also enhances the DragGAN optimization process. We employed an alternating optimization strategy, where in one iteration, we focused on modifying the latent code to initiate directional changes, followed by adjusting the heatmap centers in the subsequent iteration. 
This collaborative adjustment of both elements demonstrated a more efficient and harmonious manipulation process.

%
%

\begin{figure}[htbp]
\begin{center}
\includegraphics[width=\linewidth]{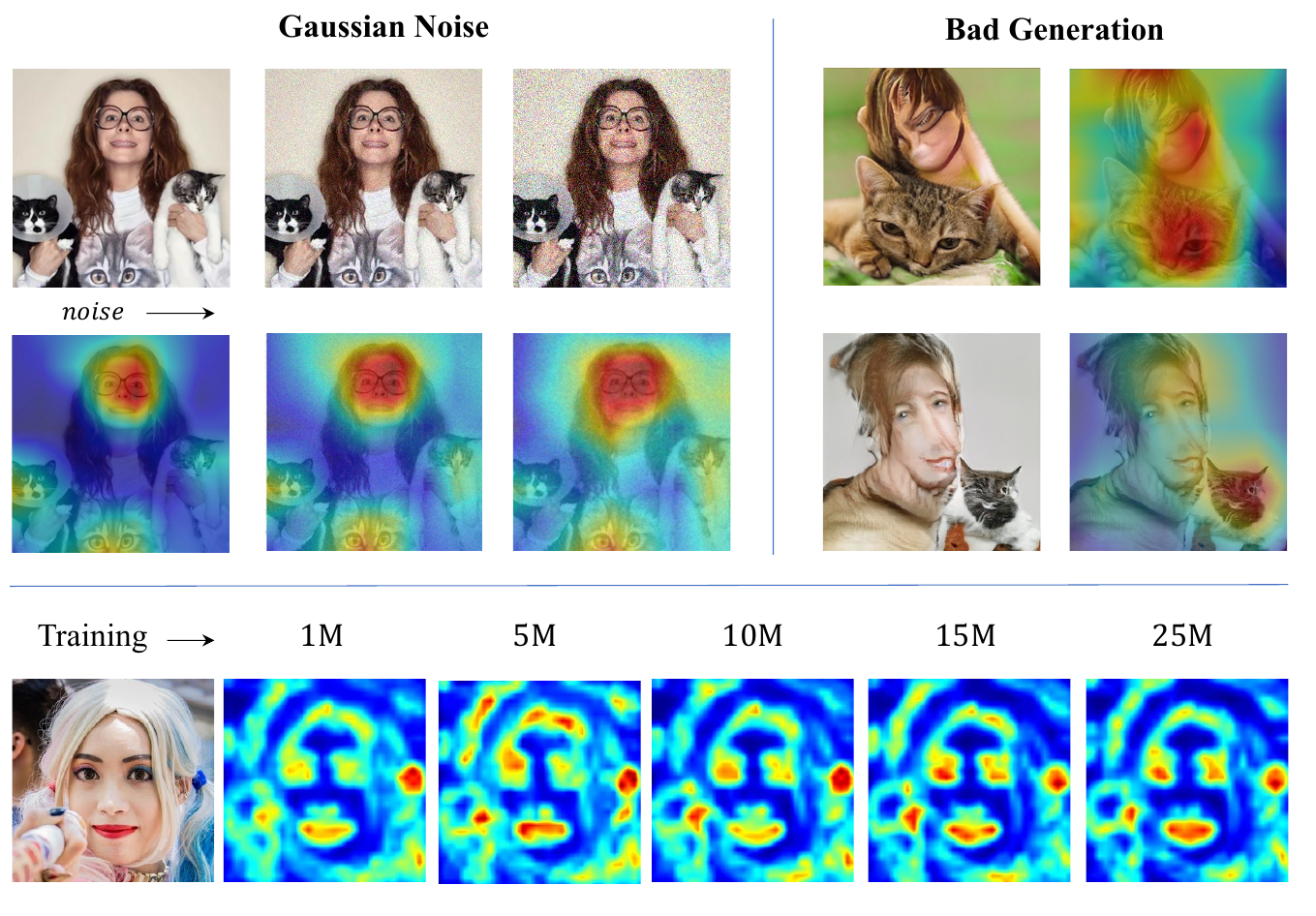}
  \caption{\textbf{Robustness and Consistency.} We test the response of $D$ to noisy images and bad generation samples in the top. The bottom visualizes that the $D$'s attention is consistent over the training. 
  }    
\label{Fig:Robustness}
\end{center}
\vspace{-10pt}
\end{figure}

\noindent \textbf{Whether Visual Attention of $D$ is Robust and Consistent?}
As discussed in ~\cref{subsec:feedback-mechanism}, the self-supervised alignment loss uses the  $D$ attention maps to guide $G$. 
It assumes the attention map from $D$ is stable enough to serve as a self-supervision signal and valid over the whole training. 
To validate the design, we first explore the \emph{robustness} of $D$. 
As shown in the left top of ~\cref{Fig:Robustness}, we add random Gaussian noise to a real image from the LSUN Cat dataset, destroying its texture. 
As the noise amplitude increasing, we can visually see the noise pattern and the local appearance has been over smoothed. 
$D$ is still attentive to the original important regions, \emph{e.g.}, the human and cat faces. 
We then test its response to terrible samples generated by a poorly-trained $G$, illustrated in the right top of ~\cref{Fig:Robustness}. 
The samples contain distorted human, cat, and background. 
That is, the visual attention of $D$ is sufficiently robust to the noise perturbation and the generated artifacts.  
Furthermore, as indicated in the bottom of ~\cref{Fig:Robustness}, we validate whether the visual attention is \emph{consistent} throughout the entire training process.  
At a very early stage of training, $D$ has already localized the discriminative regions. The focus of such visual attention is consistently maintained till the end of the training.  
The robustness and consistency property of $D$ attention could successfully provide a support for our self-supervision objective. \\
%

\begin{figure}[!ht]
    \centering
    \includegraphics[width=\linewidth]{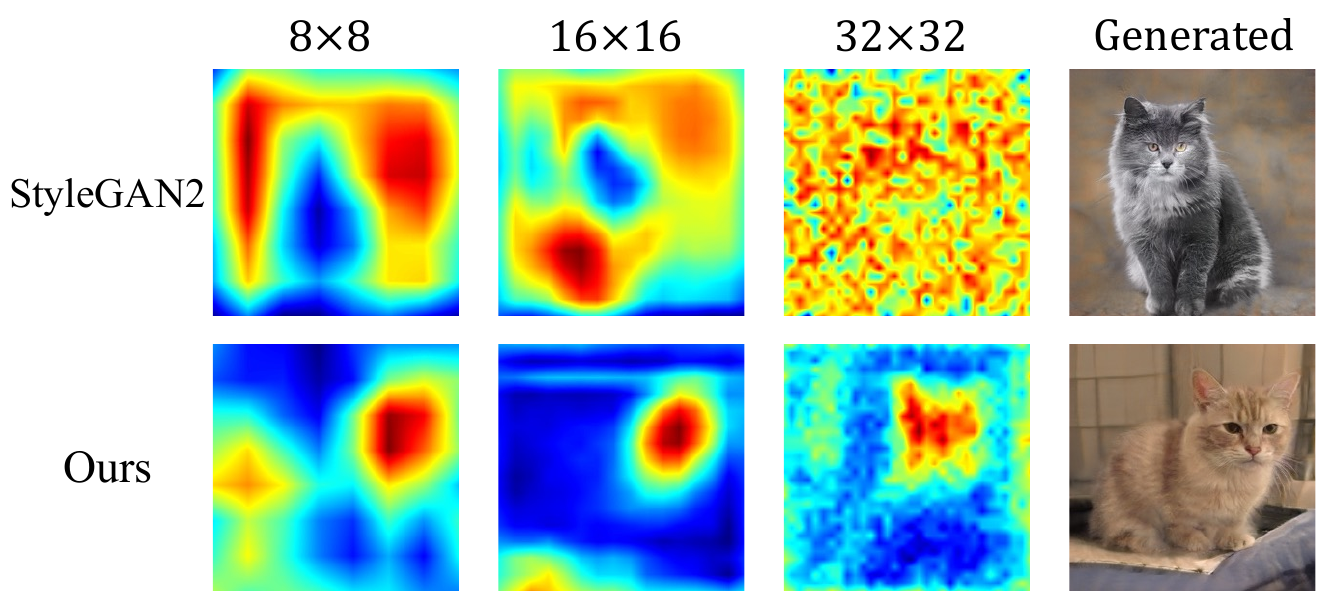}
    \caption{
        \textbf{Visualization of intermediate features of the generator}. Our method shows a much clearer spatial awareness than the baseline, whose spatial focus is close to random.
    }
    \vspace{-10pt}
    \label{Fig:Gfeat}
\end{figure}

\noindent \textbf{Visualization of Generator Intermediate Features.} 
The spatial attention of generator is worth investigating. 
However, CAM or GradCAM is not a suitable visualization tool because they both require a classification score, which is not applicable for a generator. 
Introducing another classifier may be a solution but it would involve the bias of the classifier. 
As an alternative solution, we could directly average the intermediate features of the generator along the feature dimension. 
Such a visualization can be viewed as \textit{the contribution of a layer towards certain pixels}.
As shown in \cref{Fig:Gfeat}, our generator shows a much clearer spatial awareness than the baseline which presents random spatial focus, particularly at the $32\times32$ resolution. \\

\noindent \textbf{Is a Heatmap Necessary to be Continuous?} %
As discussed in \cref{subsec:spatial-awareness-to-g}, coarse heatmap processing could improve the synthesis quality (\cref{tab:main_results}) even though it reduces the heatmap continuity, which is somehow counter-intuitive. 
We study the importance of heatmap continuity by randomly inserting extreme values ($0$ or $1$) into heatmaps, \textit{i.e.}, adding impulse noise. 
It would lead to some local jumps but keep the overall spatial structure in heatmaps.
The synthesis quality stays stable as we gradually increase the impulse noise percentage.
For example, on the LSUN Cat dataset, the model keeps an FID of $6.93$ even if using impulse noise over $5\%$ heatmap pixels.
The synthesis quality is not sensitive to the continuity of the spatial inductive bias, possibly because the real-world semantics distribution is not necessary to be continuous.
Besides, we empirically find that heatmap continuity would also not affect spatial controllability. 
\\

\noindent \textbf{How Important is the Receptive Field for Heatmap Processing?} 
Another hypothesis for the effect of coarse heatmap processing is that a global understanding matters.
A large receptive field will introduce a global view of the spatial structure.
Without coarse heatmap processing, the FID of our model on the LSUN Cat dataset is $6.81$, with the convolution kernel size as $3\times3$.
Increasing the kernel size to $7\times7$ would improve the result to $6.72$.
If adopting dilated convolution~\cite{dilated} to further increase the receptive field, it would become $6.61$, closer to the result of coarse heatmap processing ($6.57$).
The experiments verify our speculation that a global understanding of spatial structure helps.
For simplicity, we do not use dilated convolution in our method.

\section{Conclusion and Discussions}
\label{sec:con}
In this paper, we propose a method to improve the spatial steerability and synthesis quality of GANs.
Specifically, we notice that $D$ spontaneously learns its visual attention, which can serve as a self-supervision signal to raise the spatial awareness of $G$. 
%
Therefore, we encode spatial heatmaps into the intermediate features of $G$, and align the heatmaps and the attention of $D$ during training.
Qualitative results show that our method successfully makes $G$ concentrate on specific regions.
This method enables multiple spatial manipulations like moving or removing objects in the synthesis by altering the encoded heatmaps, and consistently improves the synthesis quality on various datasets.
%
\\
%

\noindent \textbf{Limitation.}
Though simple and effective, our SpatialGAN is heuristic and built upon existing techniques. 
In addition, we notice the spatial encoding operation would sometimes lead to a synthesis blurring at the location of heatmaps boundaries, which may visually affect the manipulation quality.
Sometimes, altering one sub-heatmap would change the appearance of some remote areas, which is not desired by our design.
We consider SpatialGAN as an empirical study and hope it can inspire more work on improving the image synthesis quality and controllability of GANs. 
%
\\

\noindent \textbf{Ethical Consideration.}
This paper focuses on improving the spatial controllability of GANs.
Although only using public datasets for research and following their licences, the abuse of our method may bring negative impacts through deep fake generation.
Such risks would increase as the synthesis results of GANs are becoming more and more realistic.
From the perspective of academia, these risks may be mitigated by promoting the research on deep fake detection.
It also requires the management on the models trained with sensitive data. \\
%
%


\ifCLASSOPTIONcaptionsoff
  \newpage
\fi
\bibliographystyle{IEEEtran}
\bibliography{ref}


\begin{IEEEbiography}[{\includegraphics[width=1in,height=1.25in,clip,keepaspectratio]{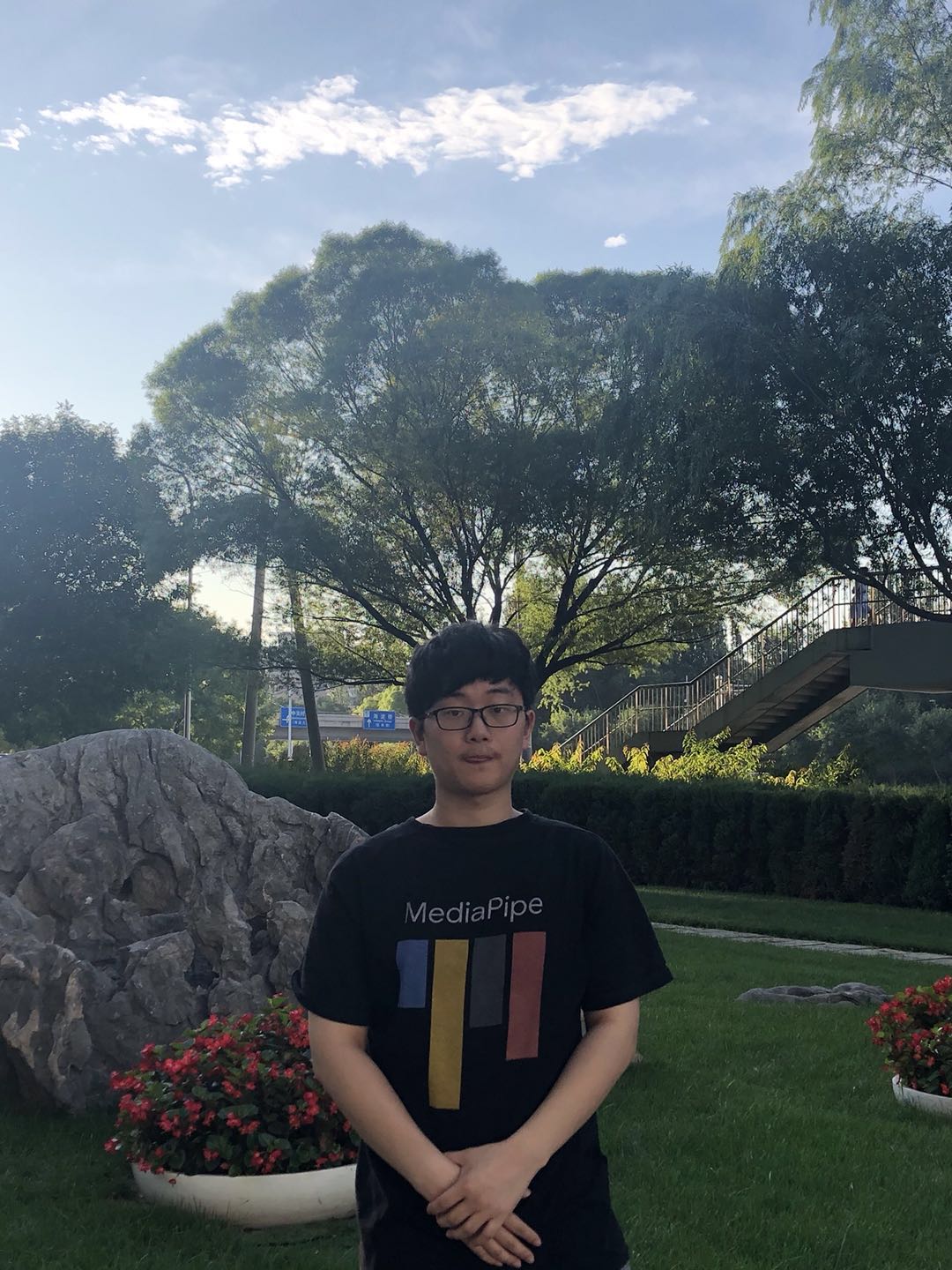}}]{Jianyuan Wang} achieved his B.E. degree with first-class honors from the Australian National University in 2019. He serves as the reviewer for TPAMI, ICLR, CVPR, ICCV, and so on. His research interests include visual geometry and generative modelling.
\end{IEEEbiography}

\begin{IEEEbiography}[{\includegraphics[width=1in,height=1.25in,clip,keepaspectratio]{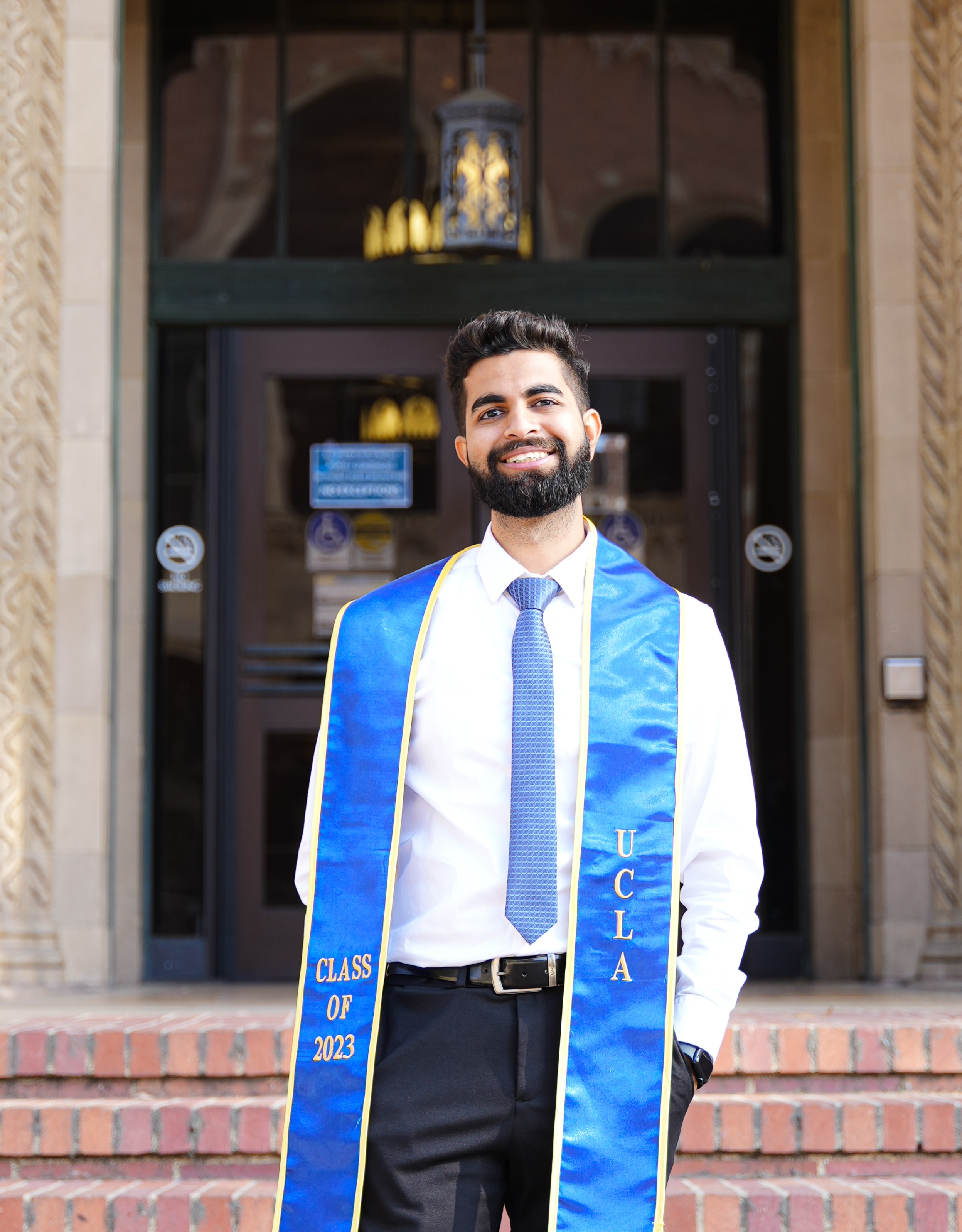}}]{Lalit Bhagat}  is pursuing his MS degree in Computer Science at the University of California, Los Angeles (UCLA). He received his B. Tech. degree in Computer Science and Engineering with honors from the JIIT, Noida, India in 2021. He is an award recipient of the GCD Fellowship at UCLA. His research interests include computer vision and generative modelling.
\end{IEEEbiography}

\begin{IEEEbiography}[{\includegraphics[width=1in,height=1.25in,clip,keepaspectratio]{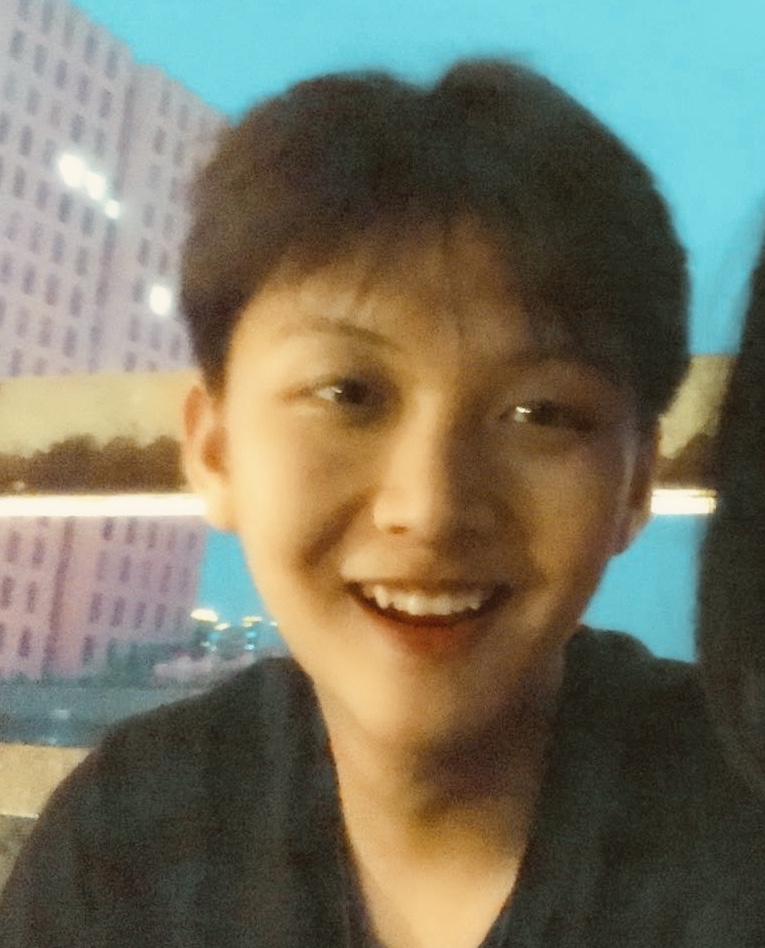}}]{Ceyuan Yang} is a young researcher at Shanghai AI Laboratory. Before that, He received his Ph.D. degree at The Chinese University of Hong Kong in 2022 and B.Eng. degree from Honors College at Northwestern Polytechnical University in 2018. His research interests include computer vision and machine learning, particularly in video understanding, generative models and representation learning.
\end{IEEEbiography}

\begin{IEEEbiography}[{\includegraphics[width=1in,height=1.25in,clip,keepaspectratio]{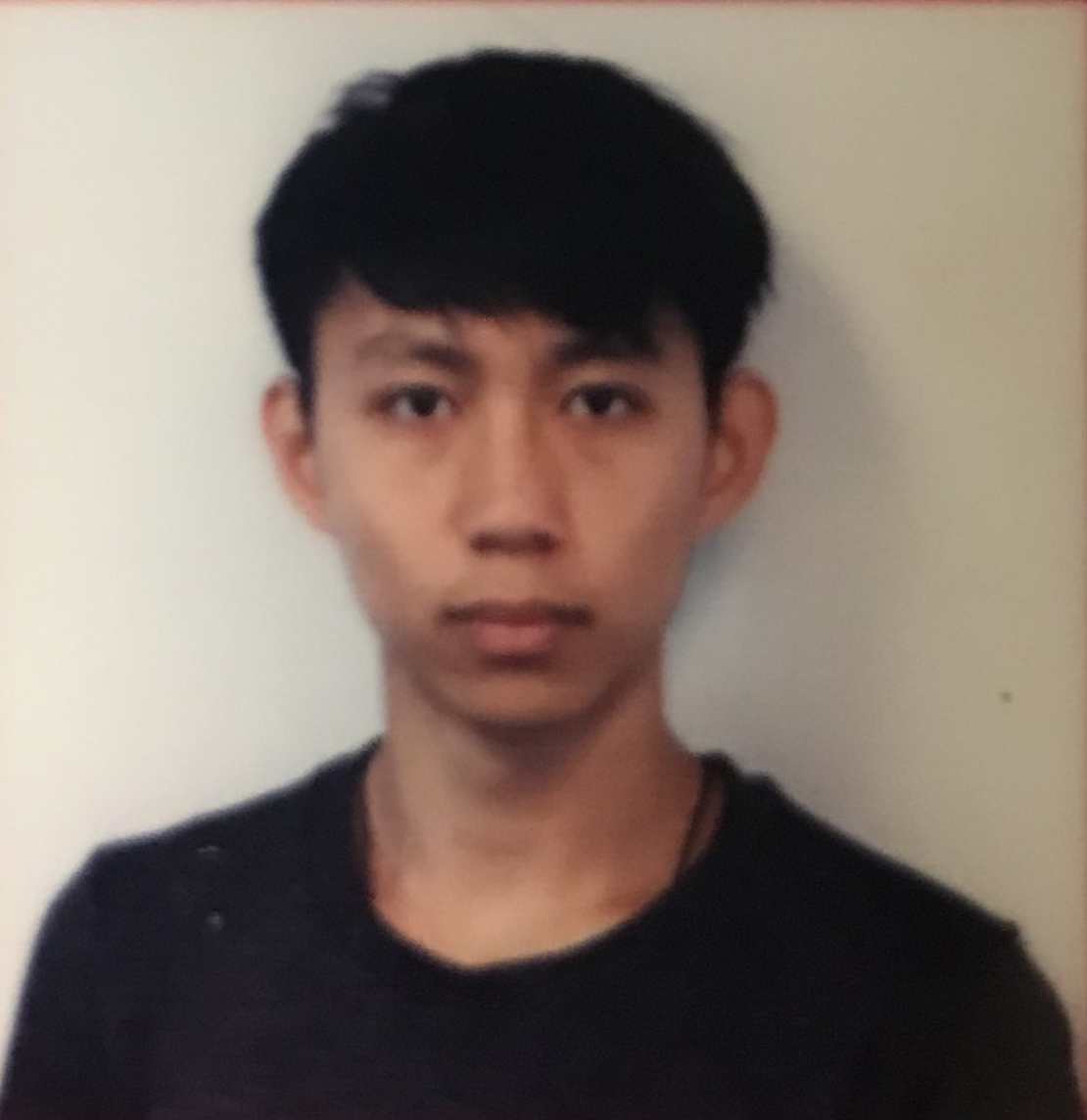}}]{Yinghao Xu}
is a third-year Ph.D student at Multimedia Lab (MMLab), Department of Information Engineering in The Chinese University of Hong Kong. His supervisor is Prof. Bolei Zhou. He graduated from Zhejiang University in 2019 working closely with Dr. Lechao Cheng. His research interests include video understanding, generative models as well as structural representation for vision perception.
\end{IEEEbiography}

\begin{IEEEbiography}[{\includegraphics[width=1in,height=1.25in,clip,keepaspectratio]{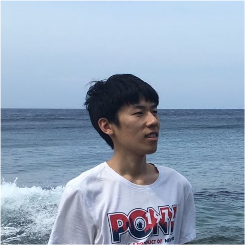}}]{Yujun Shen}
is a senior research scientist at Ant Research. Before that, he worked as a senior researcher at ByteDance Inc. He received his Ph.D. degree at the Chinese University of Hong Kong and his B.S. degree at Tsinghua University. His research interests include computer vision and deep learning, particularly in 3D vision and generative models. He is an award recipient of Hong Kong PhD Fellowship.
\end{IEEEbiography}

\begin{IEEEbiography}[{\includegraphics[width=1in,height=1.25in,clip,keepaspectratio]{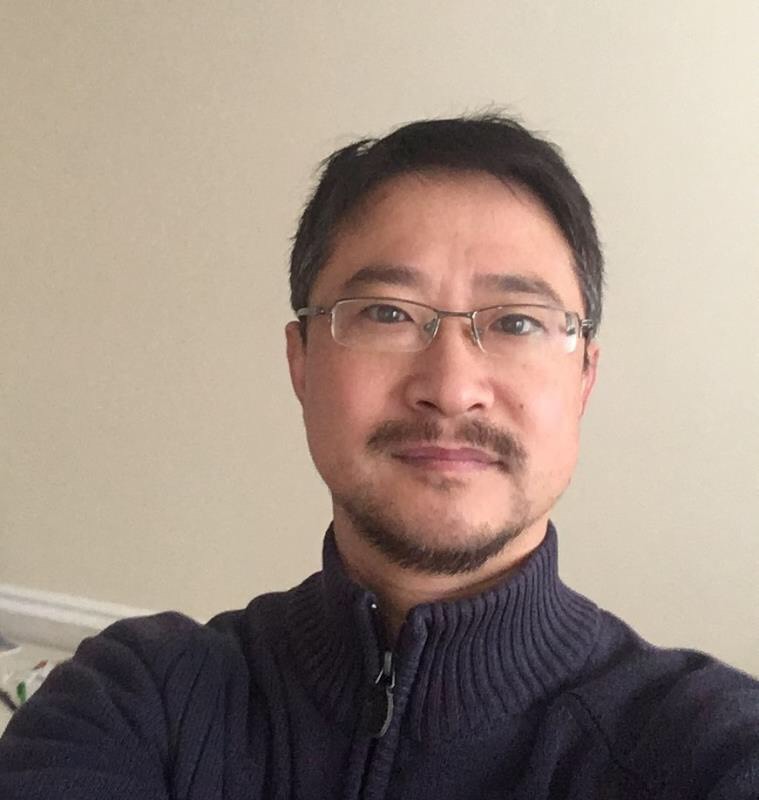}}]{Hongdong Li} is a Professor of the Australian National University (ANU). He was a visiting professor with the Robotics Institute, Carnegie Mellon University (CMU) in 2017. His research interests include 3D vision reconstruction, structure from motion, multi-view geometry, and visual perception. He is an Associate Editor for IEEE TPAMI, and serves as an Area Chair for recent years’ CVPR, ICCV, and ECCV. He is also a General Co-Chair for ACCV 2018 and ACCV 2022. He won a number of paper awards in computer vision and pattern recognition, including the CVPR Best Paper Award 2012, the Marr Prize (Honorable Mention) at ICCV 2017, and so on. 
\end{IEEEbiography}

\begin{IEEEbiography}[{\includegraphics[width=1in,height=1.25in,clip,keepaspectratio]{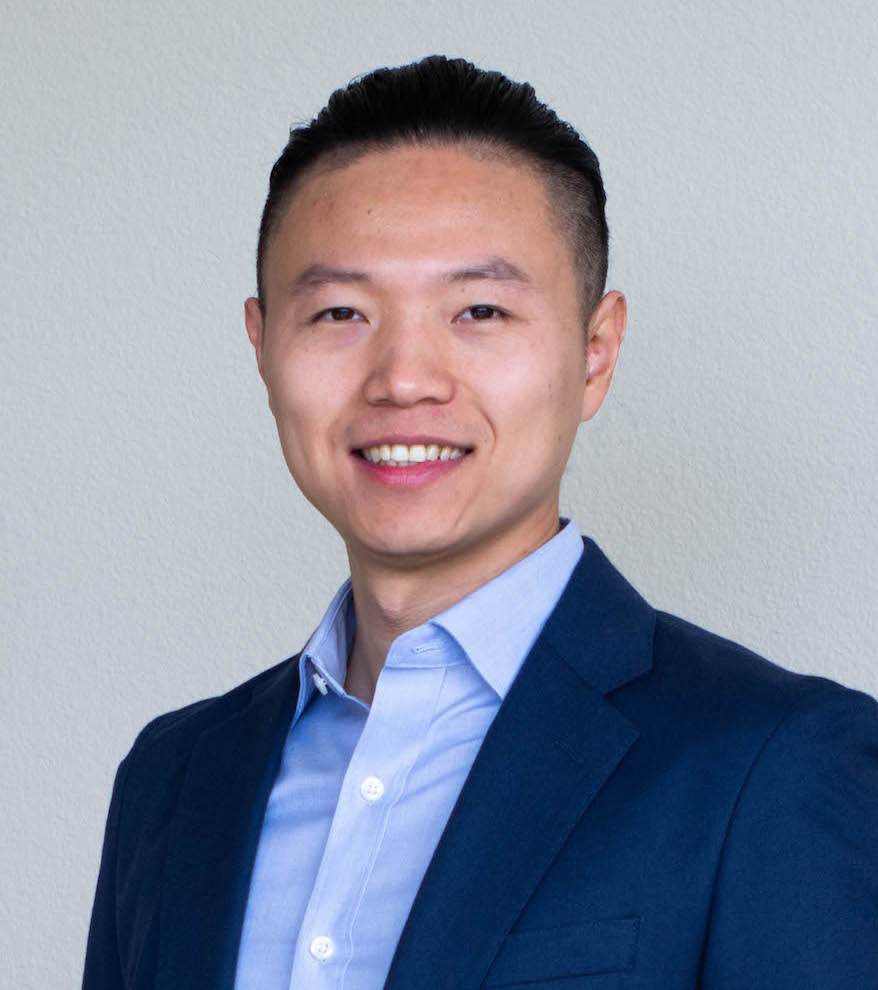}}]{Bolei Zhou} is an Assistant Professor in the Computer Science Department at the University of California, Los Angeles (UCLA). He earned his Ph.D. from MIT in 2018. His research interest lies at the intersection of computer vision and machine autonomy, focusing on enabling interpretable human-AI interaction. He and his colleagues have developed a number of widely used interpretation methods such as CAM and Network Dissection, as well as computer vision benchmarks Places and ADE20K. He is an associate editor for Pattern Recognition and has been area chair for CVPR, ICCV, ECCV, and AAAI. He received MIT Tech Review's Innovators under 35 in Asia-Pacific Award.
\end{IEEEbiography}

\end{document}



\newcommand{\titlename}{Spatial Steerability of GANs via Self-Supervision from Discriminator \\
-- \textit{Supplemental Material} --}
\title{\titlename}



\author{
Jianyuan Wang,
Ceyuan Yang, 
Yinghao Xu,
Yujun Shen,
Hongdong Li,
and Bolei Zhou\\
}


\IEEEtitleabstractindextext{


}

\maketitle
\IEEEdisplaynontitleabstractindextext
\IEEEpeerreviewmaketitle


\section*{Appendix}

As discussed in the paper, our SpatialGAN enables the interactive spatial editing of the output image. We build an interactive interface to visualize the property of spatially moving, which is illustrated in the video \textcolor{magenta}{demo.mp4}. In the appendix, we provide the following content. \cref{sec:imple_supp} includes the codes and  implementation details. \cref{sec:demo} provides a description of the demo video, and  
\cref{sec:ab_supp} provides more ablation studies on our components. \cref{sec:disscussion_supp} consists of some further discussions.

\section{Implementation}
\label{sec:imple_supp}

\noindent \textbf{Training.} We run the experiments on a computing cluster using an environment of PyTorch $1.8.1$ and CUDA $9.0$. For the convenience of reproducibility, we use the official PyTorch implementation of \href{https://github.com/NVlabs/stylegan2-ada-pytorch}{StyleGAN2} as our codebase. All the experiments follow the configuration of `paper256' in the codebase. Specifically, we use the Adam optimizer with a learning rate of $2.5\times10^{-3}$. The minibatch size is $64$ and the group size for the minibatch standard deviation layer is $8$. The depth of the mapping network is $8$. For all the datasets, we set the $R_1$ regularization weight $\gamma$ as $1$. We also adopt mixed-precision training for speedup. \\


\noindent \textbf{PyTorch-Style Codes.} We provide the PyTorch implementation for hierarchical heatmap sampling and SEL$_{\textit{norm}}$, as shown in Code~\ref{alg:code_hm_sampling} and Code~\ref{alg:code_hm_encode}. The technical description of Code~\ref{alg:code_hm_sampling} can be found in Section 3.2 of the paper. The normalization operation is conducted as $2D$ instance  normalization. The implementation of multi-object heatmap sampling is close to Code~\ref{alg:code_hm_sampling}, but has multiple centers in level $0$ and adopts the level $0$ heatmap as the one for level $1$ and $2$.\\

\noindent \textbf{Architecture of SEL$_{\textit{concat}}$.} Same as its counterpart SEL$_{\textit{norm}}$, SEL$_{\textit{concat}}$ first uses a convolutional layer to extract features from the input heatmap, with a dimension of $64$. It then concatenates the extracted features with the input feature map. Two convolutional layers are used after concatenation, with an intermediate dimension of $256$. Similarly, SEL$_{\textit{concat}}$ adopts 
a residual connection, and employs another convolutional layer for post-processing. All the convolutional layers use a kernel size of $3\times3$.

\floatname{algorithm}{Code}
\begin{algorithm}[t]
\caption{Hierarchical Heatmap Sampling in PyTorch.}
\label{alg:code_hm_sampling}
\definecolor{codeblue}{rgb}{0.25,0.5,0.5}
\lstset{
  backgroundcolor=\color{white},
  basicstyle=\fontsize{7.2pt}{7.2pt}\ttfamily\selectfont,
  columns=fullflexible,
  breaklines=true,
  captionpos=b,
  commentstyle=\fontsize{7.2pt}{7.2pt}\color{codeblue},
  keywordstyle=\fontsize{7.2pt}{7.2pt},
}
\begin{lstlisting}[language=python]
def generate_heatmap_hie(batch, res, var):
    # batch: batch size
    # res: the resolution of the training samples
    # var: variance, the influence area of sub-regions

    # sample the level 0 center by Gaussian
    center_x = torch.normal(mean=res/2,std=res/3,
                        size=(batch,1,1,1))
    center_y = torch.normal(mean=res/2,std=res/3,
                        size=(batch,1,1,1))
    centers = torch.cat((center_x,center_y),dim=1)

    # drop the sampling if the level 0 center is 
    # outside the image
    if (centers>res).any() or (centers<0).any():
        return False

    # normalize sampled centers to the range of [-1,1]
    centers = 2 * centers / (res - 1) - 1

    # build xy grids, reshape to (res, res), normalize
    x_grids = 2 * torch.arange(res).unsqueeze(0).expand 
                        (res, res) / (res - 1) - 1
    y_grids = 2 * torch.arange(res).unsqueeze(1).expand
                        (res, res) / (res - 1) - 1
    grids = torch.stack((y_grids,x_grids), dim=0).
                        unsqueeze(0)

    # generate level 0 spatial heatmap by Gaussian
    # drop the constant term to keep heatmap in (0,1)
    heatmap = {}
    heatmap['l0'] = torch.exp(-torch.square(grids -
                        centers).sum(dim=1) / var) 

    # heuristically use two and four sub-regions 
    # in level 1 and 2
    num_l1 = 2; num_l2 = 4

    # we sample level 1 and 2 centers based on the 
    # level 0 center
    l1_delta = torch.normal(mean=0,std=res/6,size=(
                        batch,2,1,num_l1))
    l2_delta = torch.normal(mean=0,std=res/6,size=(
                        batch,2,1,num_l2))

    # (b, 2, 1, n), where 2 denotes xy dim
    l1_centers = l1_delta + centers
    l2_centers = l2_delta + centers

    # decrease the influence area level by level
    var_l1 = var/np.sqrt(2); var_l2 = var_l1/np.sqrt(2)

    # (b, h, w, n)
    dis=grids.unsqueeze(-1) - l1_centers.unsqueeze(-2)
    heatmap['l1']  = torch.exp(-torch.square(dis).sum(
                        dim=1) / var_l1) 
                        
    dis=grids.unsqueeze(-1) - l2_centers.unsqueeze(-2)
    heatmap['l2'] = torch.exp(-torch.square(dis).sum(
                        dim=1) / var_l2) 
    
    return heatmap
\end{lstlisting}
\end{algorithm}

\floatname{algorithm}{Code}
\begin{algorithm}[t]
\caption{SEL$_{\textit{norm}}$ in PyTorch.}
\label{alg:code_hm_encode}
\definecolor{codeblue}{rgb}{0.25,0.5,0.5}
\lstset{
  backgroundcolor=\color{white},
  basicstyle=\fontsize{7.2pt}{7.2pt}\ttfamily\selectfont,
  columns=fullflexible,
  breaklines=true,
  captionpos=b,
  commentstyle=\fontsize{7.2pt}{7.2pt}\color{codeblue},
  keywordstyle=\fontsize{7.2pt}{7.2pt},
}
\begin{lstlisting}[language=python]
class SEL_Norm(torch.nn.Module):
    def __init__(self, dim, hm_dim, inter_dim = 64):
        super().__init__()
        # dim: dimension of the input feature map
        # hm_dim: dimension of the input spatial heatmap
        # inter_dim: intermediate dimension 

        ks = 3 # kernel size 

        # normalization module
        self.norm = nn.InstanceNorm2d(dim, affine=False)

        # layer to extract heatmap features 
        self.conv0 = Conv2dLayer(hm_dim, inter_dim, 
                        kernel_size=ks, activation='lrelu')

        # learnable functions,
        # each equipped with a convolutional layer
        self.sigma_func = Conv2dLayer(inter_dim, 1, ks)
        self.mu_func = Conv2dLayer(inter_dim, 1, ks)

        # to process the feature map 
        # after residual connection
        self.output_conv = nn.Sequential(
                    nn.LeakyReLU(2e-1),
                    Conv2dLayer(dim, dim, kernel_size=ks),)


    def forward(self, x, hm):
        # x: the input feature map 
        # hm: the input spatial heatmap

        x_s = x

        # resize heatmap to match feature map size
        hm = F.interpolate(hm, size=x.size()[2:], 
                    mode='bilinear', align_corners=True)

        # normalization
        normalized = self.norm(x)

        # heatmap feature
        hm_feature = self.conv0(hm)

        # denormalization
        sigma = self.sigma_func(hm_feature)
        mu = self.mu_func(hm_feature)
        dx = (1 + sigma) * normalized + mu

        # residual connection
        output = x_s + self.output_conv(dx) * 0.1

        return output
\end{lstlisting}
\end{algorithm}
\section{Demo}
\label{sec:demo}
Our method supports interactive editing over the output synthesis.
For better illustration, we provide an interactive interface and show it in the demo video. 
We particularly visualize the result of hierarchical editing, \textit{i.e.}, non-indoor scenes.
Specifically, given a well-trained SpatialGAN model, the `Reset' button will randomly sample a latent code, and generate an image using the default spatial heatmaps. Users can move heatmap centers by dragging. The movement of centers updates the heatmaps in real time, shown in the second column from right. From top to down, the heatmaps correspond to $4\times4$, $8\times8$, and $16\times16$ feature resolution, \emph{i.e.}, level $0,1,2$. Once setting the heatmaps, the users can click on the button `Generate' to produce an image with the unchanged latent code and moved heatmaps. They can also click the `auto' button under `Generate', which enables automatic generation after each heatmap movement. Please note that, using hierarchical heatmap sampling, the level $1$ and level $2$ centers would automatically move with level $0$ center. 


In the demo video, we can see how the church tower reacts with heatmap moving, and how SpatialGAN tries to produce high-quality results even under some extreme cases. We also observe that controlling a level $2$ heatmap center can lead to a result like  `shaking' the ear of a cat. 

\section{Ablation Study}
\label{sec:ab_supp}

\noindent \textbf{Hyper-parameters for Hierarchical Heatmap Sampling.} As mentioned in the paper, we heuristically use two sub-heatmaps in the $8\times8$ feature resolution and four sub-heatmaps in $16\times16$. Here we provide an ablation study in ~\cref{tab:ab_hm_sample} to show this setting is effective on the LSUN Cat dataset. In addition, though other settings may not be best, they all achieve reasonable results, and a clear improvement over the baseline. It verifies that the proposed method is robust to heatmap sampling hyper-parameters. \\

\begin{table}[t]
\small
\begin{center}
\caption{\textbf{Ablation study on the hyper-parameters of hierarchical spatial heatmap sampling,} on the LSUN Cat~\cite{yu2015lsun} dataset. Heuristically, using $2$ heatmap centers in the $8\times8$ (level $1$) feature resolution and $4$ centers in the $16\times16$ (level $2$) resolution lead to a good result. The baseline does not use spatial heatmaps, denoted as `N/A'. The experiments here did not use coarse heatmap processing. }
\begin{tabular}{c|c|ccc|ccc}
\hline
&Baseline& \multicolumn{3}{c|}{Level 1} & \multicolumn{3}{c}{Level 2}     \\
\hline
Num & N/A & 1 & 2 & 4 & 2 & 4 & 8 \\
\hline
FID $\downarrow$& 8.36 & 6.93 & 6.81 & 6.97 &  6.90 & 6.81 & 7.02 \\
%
\hline
\end{tabular}
\label{tab:ab_hm_sample}
\end{center}
\vspace{-15pt}
\end{table}

\noindent \textbf{Hyper-parameters for Alignment.} In the paper, Fig. 13 provides qualitative  support for the effectiveness of $\mathcal{L}_{\textit{align}}$. Here we quantitatively explore the effect of its loss weight and truncation threshold $\tau$, illustrated in ~\cref{tab:ab_align}. On the LSUN Cat dataset, different loss weights can generally lead to a satisfactory performance, where a number of $1.0$ or $1.5$ is close to the best. Therefore, we use a loss weight of $1.0$ for the experiments using $\mathcal{L}_{\textit{align}}$. We also prove that the truncation operation is beneficial, since our spatial heatmaps cannot perfectly match the real GradCAM maps with complex structures. Truncating the samples those have been `good enough' reduces the difficulty of optimization. With the help of $\tau$, we improve the FID from $7.10$ to $6.81$. Similarly, we use $\tau=0.25$ for all the datasets. 



\begin{table}[t]
\small
\begin{center}
\caption{\textbf{Ablation study on the hyper-parameters of $\mathcal{L}_{\textit{align}}$,} on the LSUN Cat dataset. We explore the effect of the loss weight and the truncation threshold $\tau$. Overall, the alignment regularization $\mathcal{L}_{\textit{align}}$ is robust to various loss weights and $\tau$ is beneficial. The experiments here did not use coarse heatmap processing. }
\tabcolsep=0.3cm
\begin{tabular}{c|c|c|c|c|c}
\hline
Loss Weight & 0.25 & 0.50 & 1.00 & 1.50 & 2.00  \\
\hline
FID $\downarrow$ & 6.99 & 6.88 &6.81 & 6.79  & 6.83 \\
%
\hline
Threshold $\tau$ & 0.00 & 0.10 & 0.25 & 0.35 & 0.50  \\
\hline
FID $\downarrow$ & 7.10 & 6.93 & 6.81 & 6.82  & 6.87 \\
%
\hline
\end{tabular}
\label{tab:ab_align}
\end{center}
\vspace{-15pt}
\end{table}


\section{Discussion}
\label{sec:disscussion_supp}
\begin{figure*}[htbp]
    \centering
    \includegraphics[width=\linewidth]{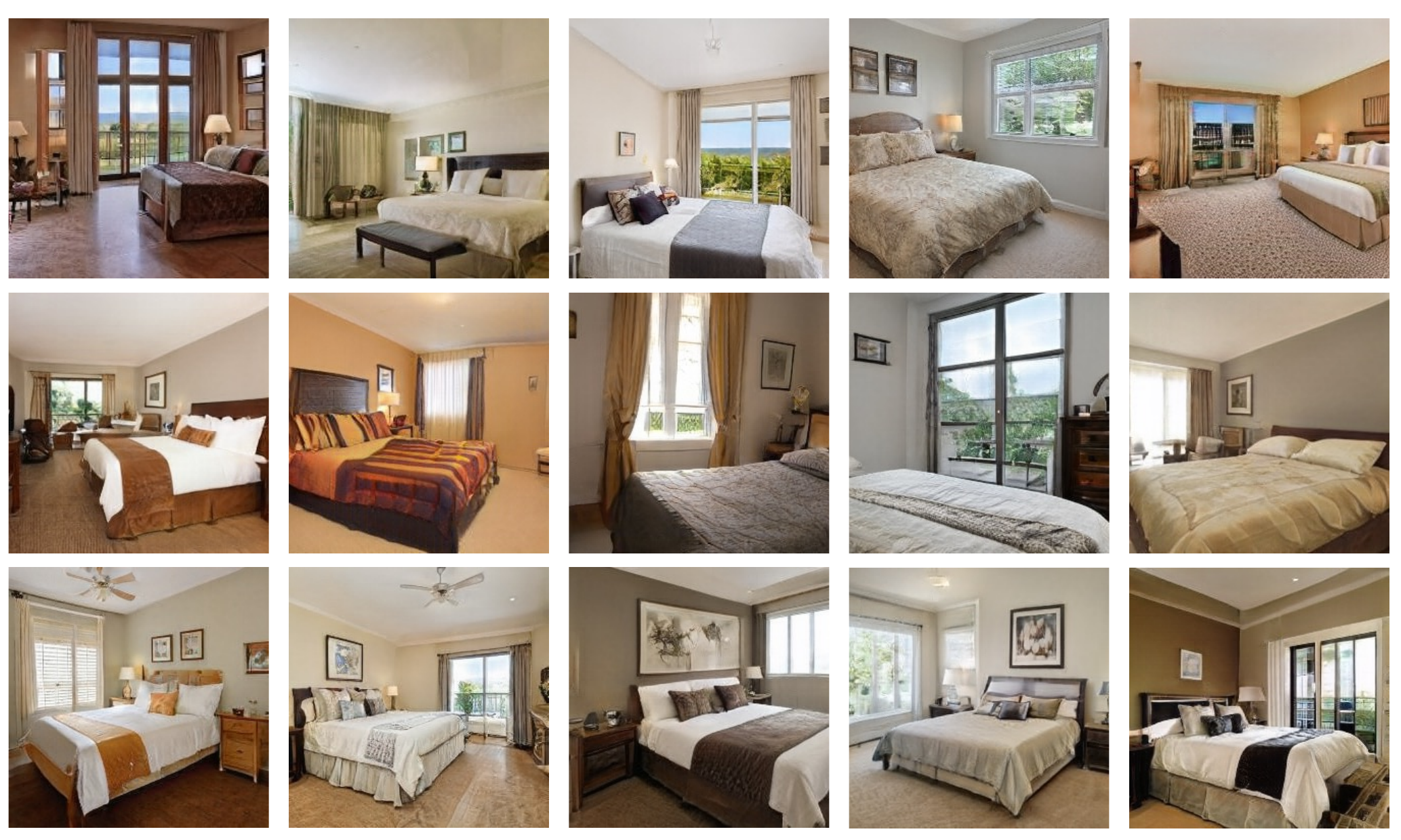}
    \vspace{-8pt}
    \caption{\textbf{Qualitative results on the LSUN Bedroom dataset.} The scenes cover different objects, layouts, and appearance styles, which show the synthesis ability and diversity of our model.
    }
    \label{Fig:Bed_Qual}
    \vspace{-5pt}
\end{figure*}

\noindent \textbf{Dataset Limitation for Manipulation.} We notice that the distribution of the training dataset limits the manipulation result. For example, the face images in the FFHQ~\cite{stylegan} dataset have been well-aligned, \emph{i.e.}, the location of face is constrained to a vertical range (close to center). Therefore, moving heatmaps to top cannot lead to an image with the face at the top. \\

\noindent \textbf{Artifacts during Heatmap Movement.} We notice that there are artifacts when interpolating spatial heatmaps, \emph{e.g.}, blurring at the location of heatmaps boundaries. We attribute this to the instability from the outputs of spatial encoding layers. Such an instability may be mitigated via involving a regularization in the way of path length regularization~\cite{stylegan2}, \emph{i.e.}, requiring a fixed-size step of heatmap movement to have a fixed-magnitude change in the image. We plan to solve these artifacts in the future work, which may further improve our synthesis quality. \\


\noindent \textbf{Synthesis Diversity.} We also provide a qualitative sample in \cref{Fig:Bed_Qual} to show the diversity of our synthesis samples. The model was trained on the LSUN Bedroom dataset. These different bedroom samples contain various objects, layouts, and appearance styles (light or dark, modern or medieval, and so on) with high quality. They verify the synthesis ability of our method.





\ifCLASSOPTIONcaptionsoff
  \newpage
\fi
\bibliographystyle{IEEEtran}
\bibliography{ref}